%% file: main.tex
\documentclass[11pt]{article}
\usepackage{blindtext}
\usepackage[utf8]{inputenc}
\usepackage{graphicx}
\usepackage{listings}
\usepackage[section]{placeins}
\usepackage[margin=1.0in]{geometry}
\usepackage{lmodern}
\usepackage{xcolor} 

\usepackage{float}
\usepackage{url}
\expandafter\def\expandafter\UrlBreaks\expandafter{\UrlBreaks
  \do\a\do\b\do\c\do\d\do\e\do\f\do\g\do\h\do\i\do\j%
  \do\k\do\l\do\m\do\n\do\o\do\p\do\q\do\r\do\s\do\t%
  \do\u\do\v\do\w\do\x\do\y\do\z\do\A\do\B\do\C\do\D%
  \do\E\do\F\do\G\do\H\do\I\do\J\do\K\do\L\do\M\do\N%
  \do\O\do\P\do\Q\do\R\do\S\do\T\do\U\do\V\do\W\do\X%
  \do\Y\do\Z}
\usepackage{authblk}

\usepackage{multicol}

\usepackage{hyperref}

\usepackage{makeidx}

\graphicspath{{../pdf/}{images/}}

\usepackage{chngcntr}
\counterwithin{figure}{section}

\makeindex

\title{\textbf{2019 Evolutionary Algorithms Review}}
\author[1]{Andrew N. Sloss}
\author[2]{Steven Gustafson}
\affil[1]{Arm Inc., Bellevue}
\affil[2]{MAANA Inc., Bellevue}

\begin{document}


\maketitle


 

\begin{abstract}
\noindent
Evolutionary algorithm research and applications began over 50 years ago. Like other artificial intelligence techniques, evolutionary algorithms will likely see increased use and development due to the increased availability of computation, more robust and available open source software libraries, and the increasing demand for artificial intelligence techniques. As these techniques become more adopted and capable, it is the right time to take a perspective of their ability to integrate into society and the human processes they intend to augment. In this review, we explore a new taxonomy of  evolutionary algorithms and resulting classifications that look at five main areas: the ability to manage the control of the environment  with  limiters, the ability to explain and  repeat  the  search process, the ability to understand input and output causality within a solution, the ability to manage algorithm bias due to data or user design, and lastly, the ability to add corrective measures.  These areas are motivated by today’s pressures on industry to conform to both societies concerns and new government regulatory rules. As many reviews of evolutionary algorithms exist, after motivating this new taxonomy, we briefly classify a broad range of algorithms and identify areas of future research.
\end{abstract}

\newpage


\newpage

\clearpage
 
\pagenumbering{arabic}


%
%

\input{sections/preface.tex}
\newpage
\input{sections/introduction.tex}

\newpage
\input{sections/evolution.tex}
\newpage
\input{sections/techniques.tex}

\newpage
\input{sections/specialized.tex}
\newpage
\input{sections/domainmap.tex}
\newpage
\input{sections/challenges.tex}
\newpage
\input{sections/predictions.tex}

\newpage
\input{sections/conclusion.tex}

\input{sections/feedback.tex}
\newpage
\input{sections/references.tex}

\printindex

\end{document}

%% file: sections/preface.tex
\section{Preface}

When attempting to find a perfect combination of chemicals for a specific problem, a chemist will undertake a set of experiments. They know roughly what needs to be achieved but not necessarily how to achieve it. A chemist will create a number of experiments. Each experiment is a combination of different chemicals. Following some theoretical basis for the experiments. The experiments are played out and the promising solutions are identified and gathered together. These new chemical combinations are then used as the basis for the next round of experiments. This procedure is repeated until hopefully a satisfactory chemical combination is discovered.\\

The reason this discovery method is adopted is because the interactions between the various chemicals is too complicated and potentially unknown. This effectively makes the problem-domain too large to explore. An Evolutionary Algorithm (EA) replaces the decision making by the chemist, using evolutionary principles to explore the problem-space. EAs handle situations that are too complex to be solved with current knowledge or capability using a form of synthetic digital evolution. The exciting part is that the solutions themselves can be original, taking advantage of effects or attributes previously unknown to the problem. EAs provide a framework that can be reused across different domains, they are mostly biologically-inspired algorithms that reside as a subbranch of Artificial Intelligence (AI).\\

Using Bertrand Russell's method of defining philosophy [\ref{ref:bertrandrussell}] i.e.\textit{“as soon as definite knowledge concerning any subject becomes possible, this subject ceases to be called philosophy, and becomes a separate science”}. AI research lives in-between philosophy and science. Ideas transition from philosophical thought to applied science and truth. Within Computer Science, the AI field resides at the edge of knowledge and as such includes a distinct part which is more philosophical and another which is more rooted in science. In this review we cover one of the science aspects. AI science incorporates many areas of research e.g. Neural Networks, Bayesian Networks, Evolutionary Algorithms, Correlation, Game Theory, Planning, Vision recognition, Decision making, Natural Language Processing, etc. It is a dynamically changing list as more discoveries are made or developed. Thirdly, \textit{Machine Learning} is the engineering discipline which applies the science to a real world problem.\\

One of the overriding motivators driving Machine Learning in recent years has been the desire to replace rigid rule-based systems. A strong candidate has been emerging which is both adaptive and outcome-based. This technology relies on data-directed inputs. Jeff Bezos, CEO of Amazon, succinctly described this concept in a letter to shareholders in 2017 [\ref{ref:jeffbezos}], \textit{"Over the past decades computers have broadly automated tasks that programmers could describe with clear rules and algorithms. Modern Machine Learning techniques now allow us to do the same for tasks where describing the precise rules is much harder."}. Also Kazuo Yano, Fellow and Corporate Officer of Hitachi Ltd, said in his keynote at the 2018 \textit{Genetic and Evolutionary Computation Conference} (GECCO) [\ref{ref:GECCO2018}] that the demand for more flexibility forces us to transition from traditional rule-oriented systems to future outcome-oriented ones.\\ 

The adaptability and transition to outcome-oriented systems means, from an end user perspective, there is more uncertainly surrounding the final result. This is construed as being either real or perceived. Rule-based systems are not impervious but tend to be deterministic and understandable e.g. the most notable being the area of safety-critical systems. This uncertainty creates the notion of User Control Attributes (UCA). The UCA include the concepts of \textit{limiters} [\ref{ref:guardrails}], \textit{explainability} [\ref{ref:GDPR}], \textit{causality} [\ref{ref:cause}], \textit{fairness} [\ref{ref:Article22},\ref{ref:googlefairness}] and \textit{correction} [\ref{ref:ainow2018report}]. These attributes have seen a lot of scrutiny in recent years due to high-profile public errors, as detailed in the \textit{AI Now 2018 Report} [\ref{ref:ainow2018report}]. The report goes into the details of fairness and highlights the various procedures, transparency and accountability required for Machine Learning systems to be safely applied in real social environments. Also worth mentioning is the \textit{International Standard Organization} (ISO), which has formed a study group focusing specifically on  \textit{Trustworthiness} [\ref{ref:isotrustworthiness}]. The study will be investigating methods to improve basic trust in Machine Learning systems by exploring transparency, verify-ability, explainability, and control-ability. The study will also look at mitigation techniques. The goal is to improve overall robustness, resiliency, reliability, accuracy, safety, security and privacy; and by doing so hopefully minimize source biases. For this review we will limit the focus to research, while at the same time being cognizant of the dangers of real world deployment.\\

Control imposes a different level of thinking, where researchers are not just given a problem to solve but the solution requires a model justifying the outcome. That model has to provide the answers to the main questions: Whether the algorithm stays within limits/restrictions? Is the algorithm explainable? Can the algorithm predict beyond historical data? Does the algorithm avoid system biases or even side-step replicating human prejudices [\ref{ref:prejudice}]? And finally, can the algorithm be corrected? These attributes are not mutually exclusive and in fact intertwine with each other. We can see a trend where modern Machine Learning algorithms will be rated not only on the quality of the results but on how well they cope with the user demanded control attributes. These attributes are to be used as a basis for a new taxonomy.\\  

This is a good point to start discussing the Computer Industry. The industry itself is facing a set of new challenges and simultaneously adding new capabilities, as summarized below:

\begin{itemize}

\item \textbf{Silicon level}: groups are starting to work on the problem of mass silicon production at the 3-nanometer scale and smaller [\ref{ref:post3nm}]. This involves designing gates and transistors at a scale unimaginable 10 or even 5-years ago, using enhanced lithographic techniques and grappling with quantum tunneling issues. These unprecedented improvements have allowed other areas higher up the software stack to flourish. Unfortunately these advancements are slowing down, as the current techniques hit both physical and economic limitations. This situation is called the \textit{End of Moore's Law} (EoML) [\ref{ref:xeoml},\ref{ref:xeoml2}]. 

\item \textbf{System level}: A number of levels above the silicon lies the system-level which also has seen some impressive advancements with the world-wide web, network infrastructure and data centers that connect everyone with everyone. The scale of the system-level advancements has opened up the possibility of mimicking small parts of the human brain. One such project is called \textit{SpiNNaker} [\ref{ref:SpiNNaker}]. SpiNNaker was upgraded and switched on in November 2018, it now consists of a million interconnected ARM cores each executing \textit{Spiking Neural Networks} (SNN) [\ref{ref:SNN}]. Even with all the hardware capability, estimates suggest that it is only equivalent to about 1\% of a human brain.

\item \textbf{Software design level}: Software design, at the other end of the spectrum, has been constantly pursuing automation. The hope being that automation leads to the ability to handle more complicated problems, which in turn provides more opportunities. New programming languages, new software paradigms and higher level data-driven solutions all contribute to improving software automation.

\end{itemize}

To handle these new challenges and capabilities requires a continuously changing toolbox of techniques. EAs are one such tool and like other tools they have intrinsic advantages and disadvantages. Recently EAs have seen a resurgence of enthusiasm, which comes at an interesting time since other branches of Machine Learning become mature and crowded. This maturity forces some researchers to explore combinations of techniques. As can be seen in this review a lot of the new focus and vigor is centered upon hybrid solutions, especially important is the area of combining evolutionary techniques with \textit{Artificial Neural Networks} (ANNs).

%% file: sections/introduction.tex
\section{Introduction}

EAs are not a new subject. In fact as we look back at some of the early computing pioneers we see examples of evolutionary discovery. For example both Alan Turing [\ref{ref:alanturing}] and John von Neumann [\ref{ref:johnvonneumann}] formed ideas around Biological Automation, Biological Mathematics and Machine Learning. These forward visionaries focused on the fact that \textit{exploitative} methods could only go so far to solve difficult problems and more \textit{exploratory} methods were required. The main difference between the two techniques is that exploitative focuses on direct local knowledge to obtain a solution whereas exploratory takes effectively a more stochastic approach (leaping into the unknown).\\


\begin{figure}[H]
\begin{center}
\includegraphics[width=1.0\textwidth]{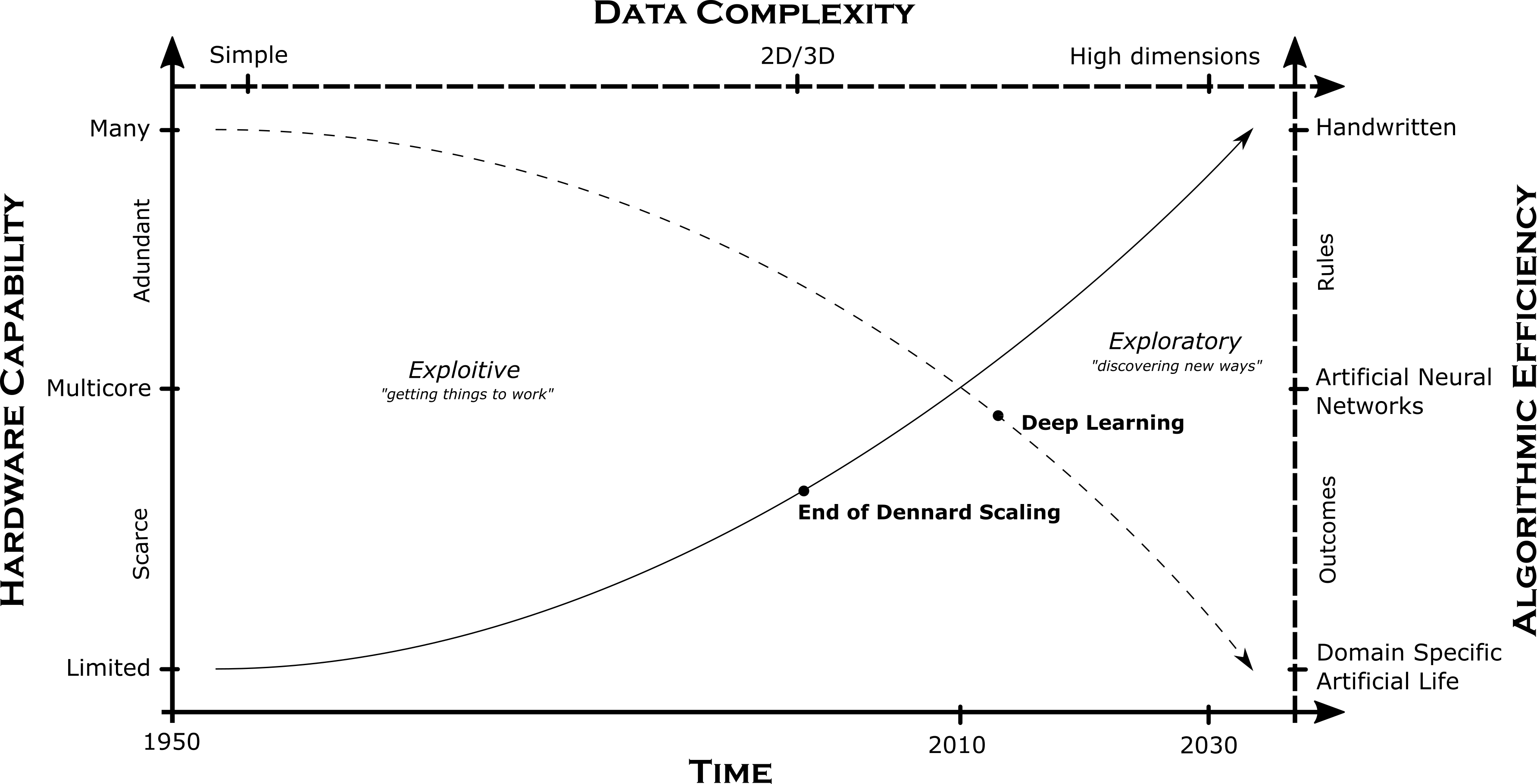}
\end{center}
\caption{Hardware capability and algorithmic efficiency over an idealized time line}
\label{figure:evolution}
\end{figure}

Figure \ref{figure:evolution} shows an idealized view of the changes to hardware capability and algorithmic efficiency over a time period. The figure shows the relationship between improvements in hardware and the types of problems that can be addressed. Before 2010, the computing industry mostly focused on exploitative problems "getting things to work efficiently". Today, due to hardware improvements, we can look at exploratory algorithms that "discover". At the extremes of the X-Y axis, top-right and bottom-right lies the respective future potentials of hardware and software i.e. the unknown.\\
\newpage

Note: \textit{End of Dennard Scaling} [\ref{ref:dennard}] marks the point in time when transistor shrinkage no-longer sustained a constant power density. In other words, static power consumption dominates the power equation for silicon. Forcing the industry to use clever frequency and design duplication techniques to mitigate the problem. \textit{Deep Learning} [\ref{ref:hinton}] represents the software resurgence of Neural Nets, due to hardware improvements and the availability of large training data sets.\\ 

The \textit{Hardware Capability} top-right of the graph represents future hardware concepts, and requires subsequent manufacturing breakthroughs. These future concepts could include alternative computing models, Quantum computing, Neuromorphic architectures, new exotic materials, Asynchronous processing, etc.. By contrast the \textit{Algorithmic Efficiency} bottom-right represents future breakthroughs in subjects like \textit{Artificial Life}, \textit{Artificial General Intelligence} (AGI), etc.; more philosophical goals than either science or engineering. Both require significant advancements beyond what we have today.\\

With these future developments, the desire is to set a problem-goal and let the "system" find the correct answer. This is extremely simple to state but highly complex to implement. And more importantly, next to impossible to implement without direct insertion of domain specific knowledge into the algorithm in question.\\

\textit{No Free Lunch Theorem} [\ref{ref:nfl}] states that no algorithm exists which outperforms every other algorithm for every problem. This means to be successful, each problem requires some form of domain specific knowledge to be efficient. The more domain specific knowledge applied to an algorithm the greater the likelihood of beating a stochastic algorithm. A stochastic algorithm can search every problem, without the requirement of domain knowledge. EAs are directed population-based stochastic search algorithms. As hardware capability increases more of these types of problems can be handled. It is the constraints of time and efficiency that forces domain knowledge to be inserted into an algorithm.\\

This paper provides an up-to-date review of the various EAs, there respective options and how they may be applied to different problem-domains. EAs are a family of biologically-inspired algorithms that take advantage of synthetic methods, namely management of populations, replication, variability and finally selection. All based upon the fundamental theory of \textit{Darwinian} evolution [\ref{ref:darwin}]. As a general rule the algorithms are often simple at a high-level but become increasingly complex as more domain knowledge is put into the system.\\

Another term frequently used to describe these style of algorithms is \textit{metaheuristics}. Metaheauristics is a higher-order concept, it is an algorithm that systematically pursues the identification of the best solution within a problem-space. EAs obviously fall under this class of algorithms and a lot of the academic literature frequently refers to metaheuristics: the main differentiator being biologically inspired algorithms.\\ 

For these algorithms to execute, some form of quantitative goal has to be set. This goal is the metric for success. The success metric can also be used as a method to exit the algorithm but most often a function of time is used. The algorithm can also be designed to be continuous i.e. never ending, always evolving. The goal itself can be made up of a single objective or multi-objectives. For multi-objective the search and optimization is towards the \textit{pareto optimal} curve i.e. attempting to satisfy all the objectives in some degree.\\

As an important side-note, EAs are mostly \textit{derivative-free}, in that majority do not require a derivative function to measure change to determine the optimal result.\\

Lastly, GECCO 2018 [\ref{ref:GECCO2018}], in Kyoto, saw a number of trends. Neuroevolution being one of the more notable ones, Neuroevolution is the method of using EAs to configure ANNs (see section \ref{section:neuroevolution}). These new trends will be tracked in more detail in future reviews if they remain popular in the evolutionary research community.

\subsection{Applications}

EAs are applied to problems where traditional exploitative or pure stochastic algorithms fail or find it difficult to reach a conclusion. This is normally due to constraints on resources, high number of dimensions or complex functionality. Solving these problems would require exceeding the available resources. In other words, given infinite resources and infinite compute capability it could be possible for a traditional exploitative or stochastic algorithm to reach a conclusion. By contrast, EAs can be thought of as the algorithms-of-last-resort. The problems in question are inherently complex, size of the problem-domain is extreme or the mere number of objectives make the problem impossibly difficult to explore. In these circumstances the solutions are more likely "good enough" solutions rather than solutions with high precision or accuracy (but this does not preclude precision or accuracy being a goal). EAs tend to be poor candidates for simple problems where standard techniques could easily be used instead. \\

They can be applied to a broad set of problem types. These types range from variable optimization problems to creating new conceptual designs. In both instances, novelty can occur which may exceed human understanding or ability. These problem-domains can be broken-down into optimization, new design and improvement.

\begin{itemize}
\item \textbf{Variable optimization} consists of searching a variable space for a "good" solution, where the number of variables being searched is large. Potentially at a magnitude greater than a traditional programming problem. This is where the goal can be strictly defined.  
\item \textbf{New structural design} consists of creating a completely new solution, for example like a program or a mechanical design. A famous example, of a non-anthropomorphized solution, is the evolutionary designed NASA antenna [\ref{ref:nasaantenna}]. The antenna design was not necessarily something a human would have created. This is where a particular outcome is desired but it is unknown how that outcome can be constructed. 
\item \textbf{Improvement} is where a known working solution is placed into the system and EAs explore potential better versions. This is where a solution already exists and there is a notion that a potential better solution can be discovered.
\end{itemize}

Being more specific on the applications side, EAs have been applied to a wide range of problems from leading edge research on self-assembly cellular automata [\ref{ref:selfassmebly}] to projecting future city landscapes for town planners (see section \ref{section:landscape}).

%% file: sections/evolution.tex
\section{Fundamentals of Digital Evolution}

Before diving directly into the fundamentals we should stress that there are two ways to describe evolution. The first is from a pure biology point-of-view dealing the various interactions of biological systems and the other is from the Computer Science perspective. In this paper we will keep the focus on Computer Science with a hint of biology. Other texts may approach the explanation from a different perspective.\\ 

\begin{figure}[H]
\begin{center}
\includegraphics[width=0.5\textwidth]{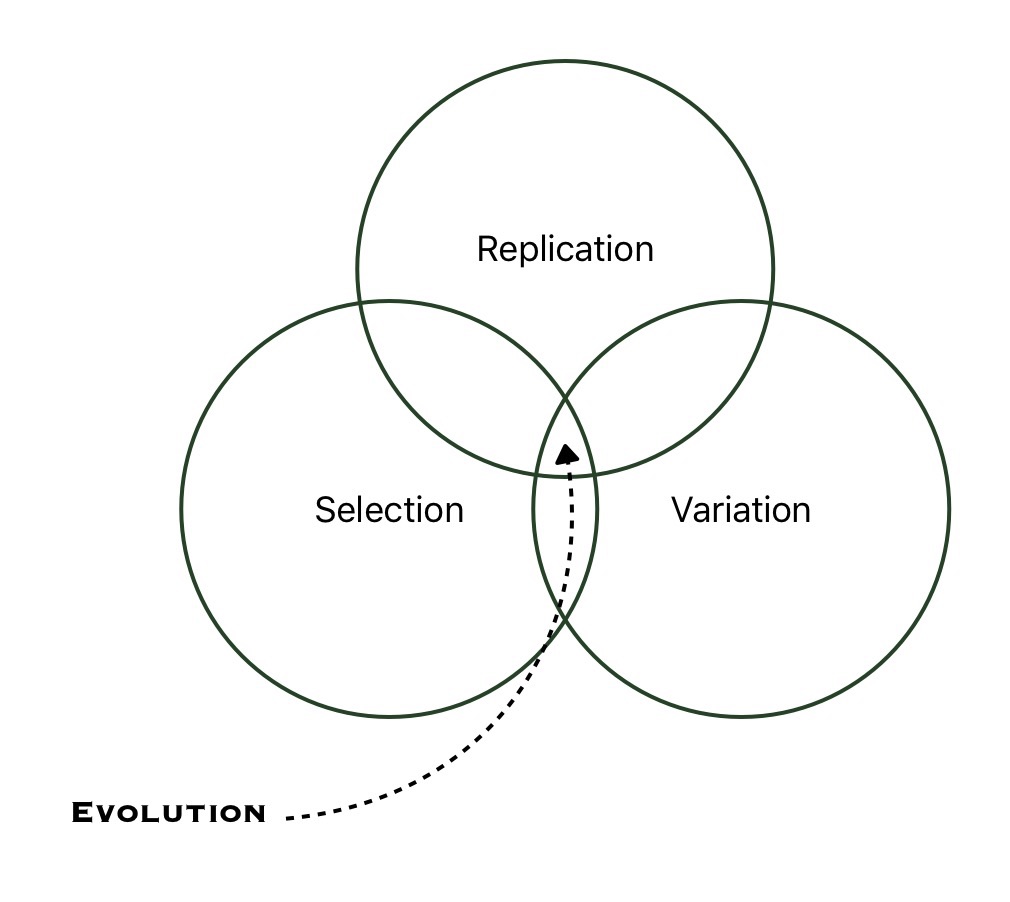}
\end{center}
\caption{Idealized Darwinian Evolution}
\label{figure:srv}
\end{figure}

Evolution is a dynamic mechanism that includes a population of entities (potential solutions) where some form of replication, variation and selection occurs on those entities, as shown in figure \ref{figure:srv}. This is a stochastic but guided process where the desire is to move towards a fixed goal. 

\begin{itemize}
\item \textbf{Replication}: Is where new entities are formed, either creating a completely new generation of the population or altering specific individuals within the same population (called \textit{steady-state}). 
\item \textbf{Variation}: Making the population diverse. There are in effect two forms of variation, namely \textit{recombination} and \textit{mutation}. Recombination, or more commonly called crossover, creates new entities by combining parts of other entities. By contrast, mutation injects randomness into the population by stochastic-ally changing specific features of the entities. 
\item \textbf{Selection}: Selection is based on Darwin's natural selection, or \textit{Survival Of The Fittest}, where selected entities that show the most promise are carried forward, with variation on the selection being used to create the children of the next generation. 
\end{itemize}

As a recap, digital evolution is one which a population of entities goes through generational changes. Each change starts with a selection from the previous generation. Each entity is evaluated against a known specific goal i.e. the fitness is established and used as input to the selection algorithm. Once a selection is made, replication occurs with different degrees of variation. The variation is either by some form of recombination from the parent selection and/or some stochastic mutation.

\begin{figure}[H]
\begin{center}
\includegraphics[width=0.7\textwidth]{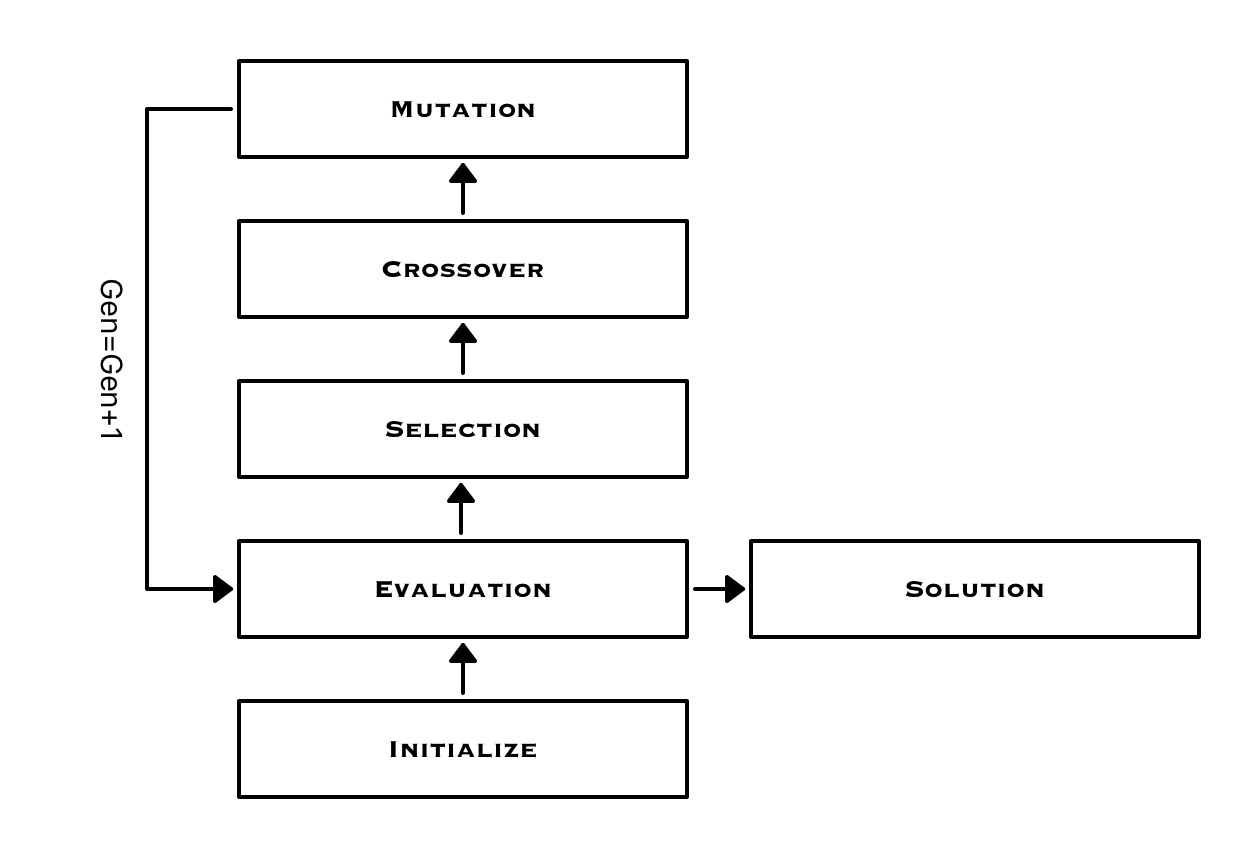}
\end{center}
\caption{Basic Digital Process}
\label{figure:process}
\end{figure}

Figure \ref{figure:process} shows the basic digital process that is adopted by many EAs. The high level synthetic evolutionary process as-shown is a relatively straightforward and simple procedure. Within the process there are many complex nuances and variations, which are influenced by our understanding of evolutionary biology. 

\subsection{Population} \label{evolution:population}

A \textit{population} is a set of solution candidates or entities. Each population is created temporally and has the potential to age with each evolutionary cycle i.e. generation. The initial population is either seeded randomly or is sampled from a set of known solutions. There are two ways for the population to be evolved. This first way consists of evolving the entire population to create the next generation. The second way is to evolve individual entities within the population (called Steady-State). Entities can die or thrive between generations.\\

The size of the population can be fixed or dynamic throughout the evolutionary process. Fixed is when the number of entities is kept constant and dynamic is when the population size changes. If dynamic, a larger initial population may bring some potential advantages. This is especially true when choosing the first strong candidates for further evolution.\\ 

A population can also be divided into subgroups, these subgroups are called \textit{demes}. A deme is a biological term that describes a group of evolving organisms with the same taxonomy. They can evolve in isolation within the population [\ref{ref:LGP}]. Demes can be made to interact with the other demes. This is often described using an island and canoe metaphor, where the demes are the islands and the interactions occur as canoes move between the islands depositing new entities. 

\subsection{Population Entities}

The population entities, or more commonly called \textit{phenotypes}, can be any type provided that type allows recombination and/or mutation to be applied. The main styles are strings and programs. In this context "programs" is a loose term. The strings are fixed length and more likely represent variables or parameters. The programs tend to be more complex. Effectively any structure from a traditional computer programming language to hardware layout, or even biological descriptions, can be a program.\\ 

Metaheuristics operate at a high-level making the process generic, allowing idea or concept can be used as the evolved substrate. Including concepts like Deep Learning Neural Network or potentially a causality equation. EAs are neutral on the subject but when it comes to specific problems, the problems themselves, tend to be strictly defined.\\ 

There are two important concepts to consider with population entities, namely \textit{niching} and \textit{crowding} [\ref{ref:simon}]. These terms are associated with diversity, which is briefly mentioned in section \ref{evolution:population}. Niching means individual entities survive generations in distinct areas of the search space. By contrast, crowding means replacing an individual entity with similar featured individuals.

\subsection{Generation} \label{section:generation}

A generation is a specific step in an evolutionary population, where individuals in the population undergo change via crossover and/or mutation. For EAs with a fixed-run a restriction is imposed on the number of generations. By contrast, continuous EAs have no upper bounds. EAs can vary from small populations with high number of generations to large populations with significantly lower number of generations. These trade-offs need to be made in terms of computation time. For example, some scenarios will have expensive evaluation functions (so fewer generations with larger populations might be preferred). The number of generations required is both algorithmic and problem-domain specific.\\ 

The children in the next generation can also include the parents from the previous generation. This is called \textit{elitism}, where the strongest entities remain in the population. 
      
\subsection{Representation and the grammar}

Representation, or more commonly called \textit{genotype}, is what EAs manipulate. There are different types of representations which include strings, tree structures, linear structures and direct-graphs. Each representation has advantages and disadvantages, and is dependent on the specific problem-domain. The representation determines what actually gets manipulated when recombination occurs, some lend themselves more to fractal/recursive like procedures and the others are more sequential and linear.\\

By contrast, the rules are defined by the grammar. EAs grammar provides the expressive boundaries for the representation. For example, in a mathematical domain adding a function like \textit{sin(x)} to the grammar provides extra richness and complexity to the representation. Similarly, for string based representations adding or removing a variable to the grammar changes the expressiveness. Other decisions can be made for structural EAs, such-as including more constructors (e.g. \textit{addition}) over destructors (e.g. \textit{subtraction}), vice-versa or more commonly keeping the grammar entirely balanced i.e. same number of constructors as destructors.\\

Possible grammars include a subset of Python or C, assembly instructions or even pure LLVM intermediate code [\ref{ref:LLVM}]. These are all potential outputs of EAs.
  
\subsection{Fitness}

Fitness is the measure of how close a result is to a desired goal. A fitness function is the algorithm used to calculate the fitness value. Calculating the fitness value for an entire population is a time consuming activity. The  time taken is related to the complexity of the fitness function and the size of the population being evaluated. The fitness function is used to select individuals for inclusion in future populations. The function can be constant throughout the evolutionary process or it can change depending upon the desired goal or situation. It is the feature of fitness function change that makes EAs highly adaptive.\\

The fitness can be calculated using different methods e.g. Area Under the Curve (AUC) from a test set, measurement of a robot responding to a set of trials, etc.; each method being problem dependent. For supervised learning it is calculated as the difference between the desired goal and the actual result obtained from the entity. Conversely, for unsupervised learning there are other methods. Once the fitness has been determined the entities can be ranked/sorted by strength. The stronger candidates are more likely to be chosen as parents for the next generation.

\subsection{Selection}

Selection is the method where individuals in the current population are chosen to be the starting parents for the next generations. In digital evolution, parents are not restricted to two; any number can be chosen. A simple method is to use the fitness value to order the population, this method is called \textit{ranked selection}. A population can be organized from the highest to lowest fitness value. The highest entities are then used as the starting parents for the next generation and so forth. As mentioned in section \ref{section:generation}, elitism is where the chosen parents remain in the next population rather than being discarded.\\ 

\textit{Diversity} [\ref{ref:simon}] is an important concept when it comes to a healthy population. Healthy populations are important for discovering "good" solutions. In other words, a diverse population has a higher exploratory capability. This tends to be important especially at the start of the search process. Diversity is directly associated with the amount of variation applied to the entities. It can be argued that local selection schemes [\ref{ref:LGP},\ref{ref:simon}] (steady-state) are naturally more likely to preserve diversity over global selection scheme. Local selection means evolution is potentially occurring at different rates across the population.\\ 

An example of a local selection scheme is called \textit{tournament selection}. As the name implies, selection requires running a tournament between a randomly chosen set of individuals. The "winner" of each tournament is then selected for further evolution.\\ 

Note: there are other selection schemes, which are not covered here.

\subsection{Multi-objective}

As the name implies \textit{multi-objective} is the concept of not having a single objective but multiple objectives. These objectives may act against each other in complex ways i.e. conflict. It is this interaction which makes multi-objective so complex. A typical example in the mobile phone industry is finding the optimum position between performance, power consumption (longevity) and cost. These three objectives can be satisfied to different degrees, effectively giving the end consumer a choice of options. Too much performance may sacrifice longevity + cost, low cost may sacrifice performance + longevity and so-on. EAs are extremely good at exploring multi-objective problems where the fitness is around a compromise along the \textit{Pareto curve}. 

\subsection{Constraints} \label{sec:constraints}

Constraints are the physical goals, as compared with the objectives which are the logical goals. The physical goals represent the real world limitations. They take a theoretical problem and make it realistic. Constraints are the limitations imposed on the entities. A constraint could be \textit{code size}, \textit{energy consumption} or \textit{execution time}. EA Researchers have discovered some of the most interesting and potentially best solutions tend to lie somewhere at the edge of the constraint boundaries.
 
\subsection{Exploitative-exploratory search} \label{exploitive-exploratorycontrol}

EAs use \textit{recombination} and \textit{mutation} for exploitative and exploratory search. The more mutation that occurs the more exploratory the search, and correspondingly the less mutation the more exploitative the search. EAs can be at either end of the spectrum, with only recombination (more exploitative) or only mutation (more exploratory). The ratios of recombination and mutation can be fixed or dynamic. By making the ratios dynamic EAs can adapt to changing circumstances. This shift may occur when the potential "good" solution is perceived to be either near or far. Another way to view this is that mutation is a local search and recombination (or crossover) is a global search. Recombination, despite using only existing genetic material, often takes much larger jumps in the search space than does mutation.

\subsection{Execution environment, modularity and system scale}

EAs can execute as a process within an Operating System, as a self-constructed dynamic program feed into a language interpreter (e.g. Python \textit{exec(open("ea.py").read())}), or within a simulator, where the simulator can be a physics simulator, biological simulator and/or a processor simulator. EAs are generic and literally any executing model can be used to explore a desired problem-domain.\\

To handle larger problems some form of modularity has to be adopted. There are many schemes including ones that introduce tree based processing or Byzantine style algorithms i.e. voting systems. Modularity tends to work best when the problem granularity is relatively small and concise, such as a function-call. The normal questions asked are (1) whether all the function-calls use the same input data, (2) whether all information is shared between the functions, or (2) whether a hierarchy of evolution has to be adopted i.e. from function call to full solution or from local to global parameter configuration.\\  

EAs can scale from one process to many processes running on several server clusters [\ref{ref:seanluke},\ref{ref:island},\ref{ref:islands2},\ref{ref:genislands}]. These compute clusters are called \textit{islands}. They  operate in either a  parallel and/or distributed fashion. A parallel system is a set of evolving islands working together towards a common goal. Genetic material or solutions can be shared. We should highlight that there are a few options when it comes to parallel topologies. By comparison, the distributed approach, which can include parallel islands, is about the physical aspect of running on various hardware systems. With both approaches, scale-out co-ordination becomes an important issue i.e. the management of the islands becomes part of the performance equation.\\

Lastly, it is important to mention \textit{co-evolution} in the context of scaling. Co-evolution is where two or more evolving populations (effectively \textit{species}) start interfering/cooperating with each other. This is particular important when digital evolution is being used to build much larger systems. Both modularity and system scale add an extra layer of complexity. Scale is a required necessity to answer more complicated problems.   

\subsection{Code bloat and clean-up}

The EAs which play with structures can quite easily have to deal with the problem of bloat [\ref{ref:codebloat}]. Bloat is a byproduct of exploring a problem-domain. Historically this was a major issue with the earlier algorithms due to hardware limitations. Today, modern systems have an abundance of compute and storage. This does not mean the limitation has gone away but it is mitigated to a certain extent. Bloat may be critically important to the evolutionary process. In nature the more that is discovered the more it seems that very little is actually wasted. Non-coding regions are not bloat, they are crucial parts to the process of molecular mechanisms.\\

There are structures or code sequences that have no value, as-in the result is circumvented by other code or structures. These neutral or noneffective structures are called \textit{introns}. Intron is a biological term referring to the noneffective fragments found in DNA. For software programs, introns are code sequences that are noneffective or neutral; these sequences can be identified and cleaned-up, i.e. eliminated. The elimination can occur either during the evolutionary process itself or at a final stage. Note, introns can be critically important to the process, so early removal can be detrimental. 

\subsection{Non-convergence, or early local optima}

There has been a lot of research focusing on the problems of non-convergence and early-local-optima solutions. Non-convergence means that the evolving entities are not making enough progress towards a solution. Early-local-optima means a sub-optimal solution has been found at the beginning of the evolutionary process. This sub-optimal solution has caused the algorithm to limit further exploration, reducing the chance of finding a better solution.\\

Non-convergence is caused by many factors including the possibility of not having the right data. EAs rely on stochastic processes to move toward, which means that the paths taken are unique, unrepeatable and non-deterministic; unique in the sense that the paths taken are always different, unrepeatable as-in randomness is used to determine the next direction, non-deterministic as-in the length of time to solution is variable. To get to a potential solution may require reruns of the algorithm. Each rerun potentially requiring some form of fine adjustment to help narrow into a good solution. In the end, when everything else fails, more domain specific knowledge may have to be inserted before convergence eventually occurs. It is important to stress that the final solution may very well be deterministic and repeatable, it is the evolutionary process to create the solution which may not be.\\ 

Similar to non-convergence is the problem of reaching a local extrema too early, and then EAs iterate persistently around a point not discovering a better solution. Again, the techniques used for non-convergence tend to be used to avoid the local optima scenario. Identification of a local optima can be difficult since the global optima is unknown. After multiple readjustments hopefully the global optima can be discovered.

\subsection{Other useful terms}

There are other terms which are worthy of a mention and brief descriptions. The first terms are the  \textit{Baldwin Effect} [\ref{baldwin}] and \textit{Lamarckian Evolution} [\ref{lamarckism}] both important concepts for digital evolution. The Baldwin Effect is about how learned behavior effects evolution. This is important for EAs that improve towards a solution under techniques such as elitism (as briefly mentioned in section \ref{section:generation}). And Lamarckian Evolution which theorizes that children can inherit characteristics from the experiences gained by their parents. Again, an important concept with direct implications for digital evolution.\\ 

By contrast, the term \textit{overfitting} [\ref{ref:overfitting}] describes a situation that should be avoided. It is when the data noise containing irrelevant information and what is actually being discovered combine into a result (configuration parameters). EAs overfit when the discovered parameters satisfies the complete data-set and not the data for the specific exploration, making it effectively useless for any future prediction using other input data sources. The potential concern is the increased risk of a \textit{false-positive} outcome.\\ 

Lastly, \textit{Genetic Drift} [\ref{ref:drift1},\ref{ref:drift2}] is a basic evolutionary mechanism found in nature, where some genotype entries between generations leave more descendants or parts than others. These descendants are in the population by random chance. They are neither in the next generation because of a strong attribute nor higher fitness value. In nature this happens to all populations and there is little chance for avoidance. EAs genetic drift can be as a result of a combination of factors, primarily related to selection, fitness function and representation. It happens by unintentional loss of genotypes. For example, random chance that a good genotype solution never gets selected for reproduction. Or, if there is a “lifespan” to a solution and it dies before it can reproduce. Normally such a genotype only resides in the population for a limited number of generations. 
 

%% file: sections/techniques.tex
\section{Traditional techniques}

In this section we briefly cover the traditional and well known EAs. These EAs tend to be older and more mature techniques. The techniques covered are frequently used by industry and research. There are numerous support frameworks available to experiment with [\ref{ref:framework-ga},\ref{ref:ponyge2},\ref{ref:pushgp},\ref{ref:framework-darwin}]. Assume for each technique discussed that there are many more variations available.

\begin{figure}[H]
\begin{center}
\includegraphics[width=0.7\textwidth]{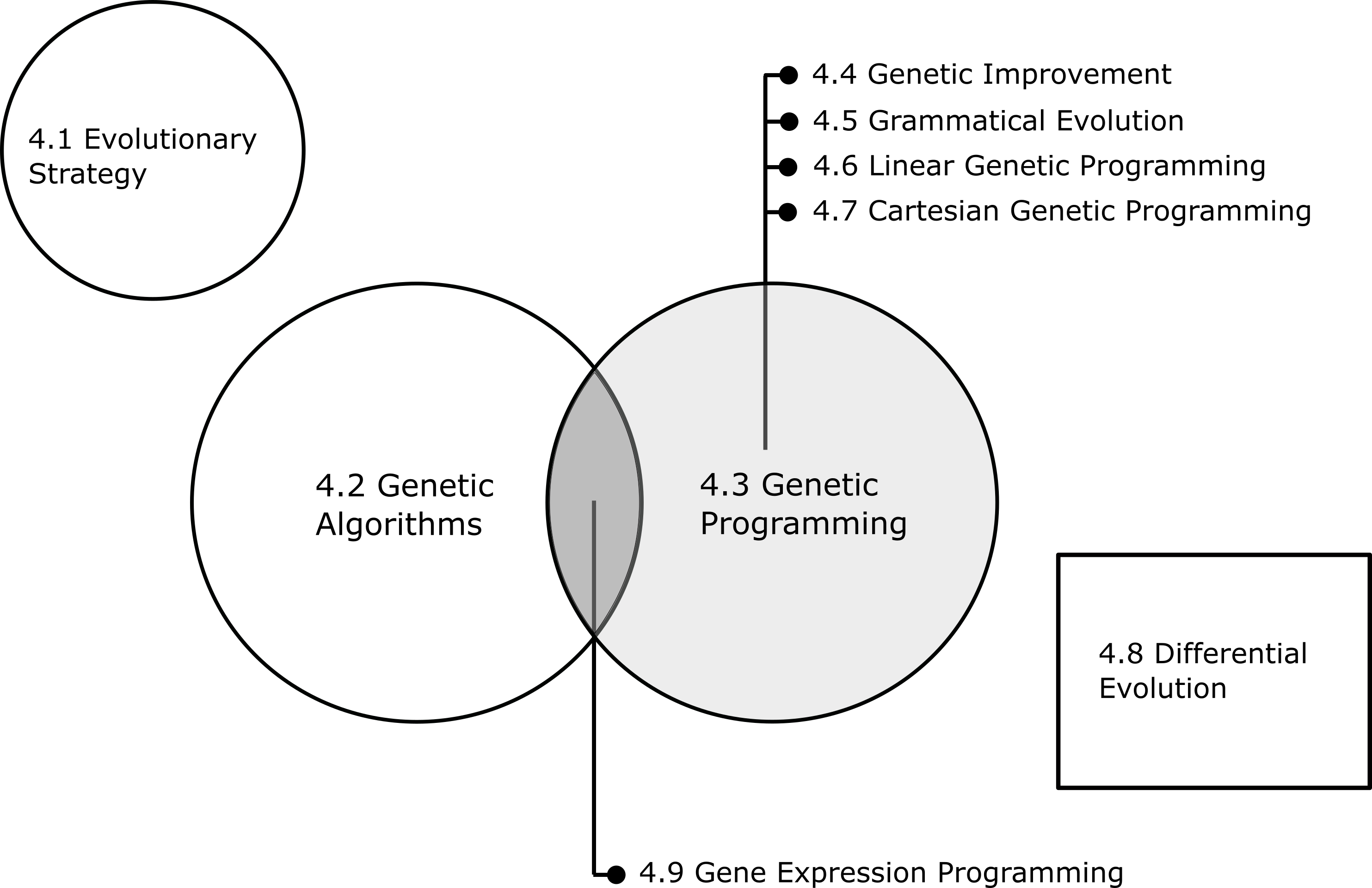}
\end{center}
\caption{Relationships between traditional EA techniques}
\label{figure:eatrad}
\end{figure}

Figure \ref{figure:eatrad} shows the relationships between the various traditional techniques. For this review we have decided to focus more on Genetic Programming and the various sub-categories. In future reviews this emphasis will likely change. 

\subsection{Evolutionary Strategy, ES} \label{evolutionarystrategy}

\textit{Evolutionary Strategy}, ES [\ref{ref:simon},\ref{ref:seanluke},\ref{ref:esrandom}] is one of the oldest EAs, developed in the 1960s at the Technical University of Berlin; it usually only involves mutation and selection. Entities are selected using \textit{truncation selection}. After the entities are evaluated, the entries below the truncation point are systematically removed from the population. The remaining parents are mutated to buildup a new population.\\

Modern implementations include Co-variance Matrix Adaptation - Evolutionary Strategy CMA-ES [\ref{cma-es}]  and an alternative Co-variance Matrix Self Adaptation - Evolutionary Strategy CMSA-ES. CMA-ES is thought to be complicated to implement, and CMSA-ES is a newer alternative and is believed to be easier to implement [\ref{ref:cmsa-es}].\\ 

\textbf{What problems do ESs solve?} An ES is used to solve continuous parameter optimization. A parameter is defined by its type and interval range (upper and lower bounds). A continuous parameter can take any value within the interval range. The precision determines the minimum change value.

\subsection{Genetic Algorithms, GA} \label{geneticalgorithms}

\textit{Genetic Algorithms}, GA [\ref{ref:simon},\ref{ref:goldberg},\ref{ref:seanluke}] is the most common and popular among the EAs. GA applies evolution to fixed length strings. The length of the string represents the dimensionality of the problem. These strings represent variables or parameters and are useful when exploring a large number problem-domain space. This space is normally beyond human or traditional methods. As well as being popular, GAs are also the most commonly taught algorithm within the various EAs. Variables or parameters are converted to fixed length strings, the strings are entities in the population and are evolved using crossover windows and mutation. Crossover is ubiquitous but explicit crossover windows are not. By comparison to an ES (section \ref{evolutionarystrategy}), a GA tends to be more generic.\\

It should be noted that a recent trend has emerged, in both GAs and ESs, where crossover is dropped and mutation is the sole mechanism for evolution (this differs to the earlier thinking expressed in the classical literature).\\ 

\textbf{What problems do GAs solve?} GAs handle optimization and configuration problems where there are too many variables or parameters for a traditional method to succeed. The variables or parameters may interact making a potential solution much harder to identify.

\subsection{Genetic Programming, GP} \label{geneticprogramming}

\textit{Genetic Programming}, GP [\ref{ref:fieldguide},\ref{ref:kovo1},\ref{ref:kovo2},\ref{ref:kovo3}] in contrast to GAs, manipulate structures and in-particular executable programs or mathematical equations. Early GPs were based on tree representations and used the \textit{LISP} programming language as the grammar. LISP was chosen for its operator richness and was relatively easily to manipulate. More recently there have been other representations introduced and newer languages such as \textit{Python} have become popular as the main target. The recombination carries out the global search, whereas the mutation covers the local search. It is frequently common for mutation to be limited to 5-10\% of the population [\ref{ref:fieldguide}]. There are always exceptions, especially if the \% mechanisms are dynamically altered between generations [\ref{ref:simon}].\\  

\textbf{What problems do GPs solve?} GPs apply evolutionary techniques to code or functions. GPs handle the manipulation of programs, so that problems that are linear, tree or direct-graph based can be explored. GPs can produce original source code, and in fact find new novel solutions to any structural style problem. In industry, GP are mostly used to discover best fit mathematical equations.

\subsection{Genetic Improvement, GI} \label{geneticimprovement}

\textit{Genetic Improvement}, GI [\ref{ref:gi}] is a subclass of GP (section \ref{geneticprogramming}), where instead of a random initial seeded population, a working program is inserted as the starting point to spawn entities of the first population. This is a powerful concept since it does not only search for a better optimized solution but also has the potential to discover and correct faults in the original "working" code. \\ 

\textbf{What problems do GIs solve?} Solves an interesting problem, where either the working code is potentially un-optimized and a more optimized version is required or bringing legacy code up to current standards.

\subsection{Grammatical Evolution, GE} \label{grammaticalevolution}

\textit{Grammatical Evolution}, GE [\ref{ref:ponyge2}] is a powerful technique. It is yet another subclass of GP (section \ref{geneticprogramming}) but instead of using a fixed grammar to evolve-able solutions, the grammar itself is select-able. A good example of GE is the \textit{PonyGE2} [\ref{ref:ponyge2}] tool, written in Python. It takes a standard Backus-Naur Form (BNF) grammar [\ref{BNF}] as an input and uses it to evolves solutions. This is a powerful method especially when dealing with more obscure programming languages. GE can also carry out GI (see section \ref{geneticimprovement}). Note the PonyGE2 source code is available on GitHub.\\

\textbf{What problems does GEs solve?} GE  solves the problem of evolving multiple programming languages using the same tool. As long as the language has a BNF-style definition, it can be evolved. This makes GE flexible across a number of problem-domains, and especially ones which require a specific programming language. 

\subsection{Linear Genetic Programming, LGP} \label{lineargeneticprogramming}

\textit{Linear Genetic Programming}, LGP [\ref{ref:LGP}] is a subclass of GP (section \ref{geneticprogramming}) and as the name implies uses a linear structure representation. The linear structure has some advantages over the more complicated tree or directed-graph structures. LGP is particularly useful for problems which are more sequential. For example, optimizing low level assembly output. It also makes the problem of manipulating complex structures easier since it is a linear flow that is being evolved. Constructs like \textit{if-style control flow} or \textit{loops} are superimposed onto the linear structure. The linear aspect of this technique introduces an ordering constraint, which potentially has \textit{Turing Machine} and/or \textit{Turing complete} ramifications.\\

\textbf{What problems do LPGs solve?} LPG solves problems that are sequential. This is particular useful for optimizing programs and low level assembly style output. Or any problem-domain where the problem being explored is about \textit{sequential ordering}.

\subsection{Cartesian Genetic Programming, CGP} \label{cartesiangeneticprogramming}

Rather than linear or tree based, \textit{Cartesian Genetic Programming} CGP [\ref{ref:CGP}] is based on Cartesian co-ordinates and directed-graphs. One basic characteristic is that the population is small (e.g. population size around 5). The small population goes through a large number of generations. CGP is uniquely qualified to handle specific problems extremely well. EAs themselves can be temporal by default. CGP introduces the concept of spatial awareness into EAs.\\

\textbf{What problems do CGPs solve?} CGP has been shown to be useful at circuit layout design since the logic components require some form of spatial awareness. Interestingly since CGP is spatial it can also be used to produce artistic designs/patterns. Recent research shows that CGP can achieve competitive results on the Atari benchmark set [\ref{ref:cgp_atari}]. CGP can also encode ANNs by adding weights to the links in the graph, allowing them to do neuroevolution (see section \ref{section:neuroevolution}).

\subsection{Differential Evolution, DE}
  
\textit{Differential Evolution}, DE [\ref{ref:simon},\ref{ref:seanluke}] is an example of a non-biologically inspired algorithm but falls under the metaheuristic category. It is based on iterating a population towards a quality goal. The iteration involves recombination, evaluation and selection. It avoids the need for gradient descent. A new candidate is based on a weighted difference between random candidates to create a third candidate, shifting the population to a higher quality level. Each new population effectively self-organizes.\\  

\textbf{What problems do DEs solves?} Works best on Boolean, Integer spaces and Reals. DE was developed specifically to find the Chebyshev polynomial coefficients and the optimization of digital filter coefficients. 

\subsection{Gene Expression Programming, GEP} \label{trad:geneexpressionprogramming}

\textit{Gene Expression Programming}, GEP [\ref{ref:gep},\ref{ref:gepferreira}] is a subclass of both GA (section \ref{geneticalgorithms}) and GP (section \ref{geneticprogramming}). This method borrows from both techniques, as-in it uses fixed length strings which encode expression trees. The expression trees can be of varied size. Evolution occurs on the simple linear, fixed length strings.\\

\textbf{What problems do GEPs solve?} It offers a powerful linear encoding which is guaranteed to produce valid programs, since GEP  follows the syntactic rules of the specific programming language being targeted. This makes it easy to implement powerful genetic operators.

%% file: sections/specialized.tex
\section{Specialized techniques and concepts}

In this section we cover some of the more exotic EAs and extended tools. These EAs are new, hybrids or just miscellaneous concepts. This is not an exhaustive list but a more holistic subset of ideas that do not follow the traditional evolutionary methods. A few of the techniques covered in this section are not technically based on biological or synthetic evolution but play an important role in the process or are placed here due to taxonomy convenience. 

\begin{figure}[H]
\begin{center}
\includegraphics[width=0.7\textwidth]{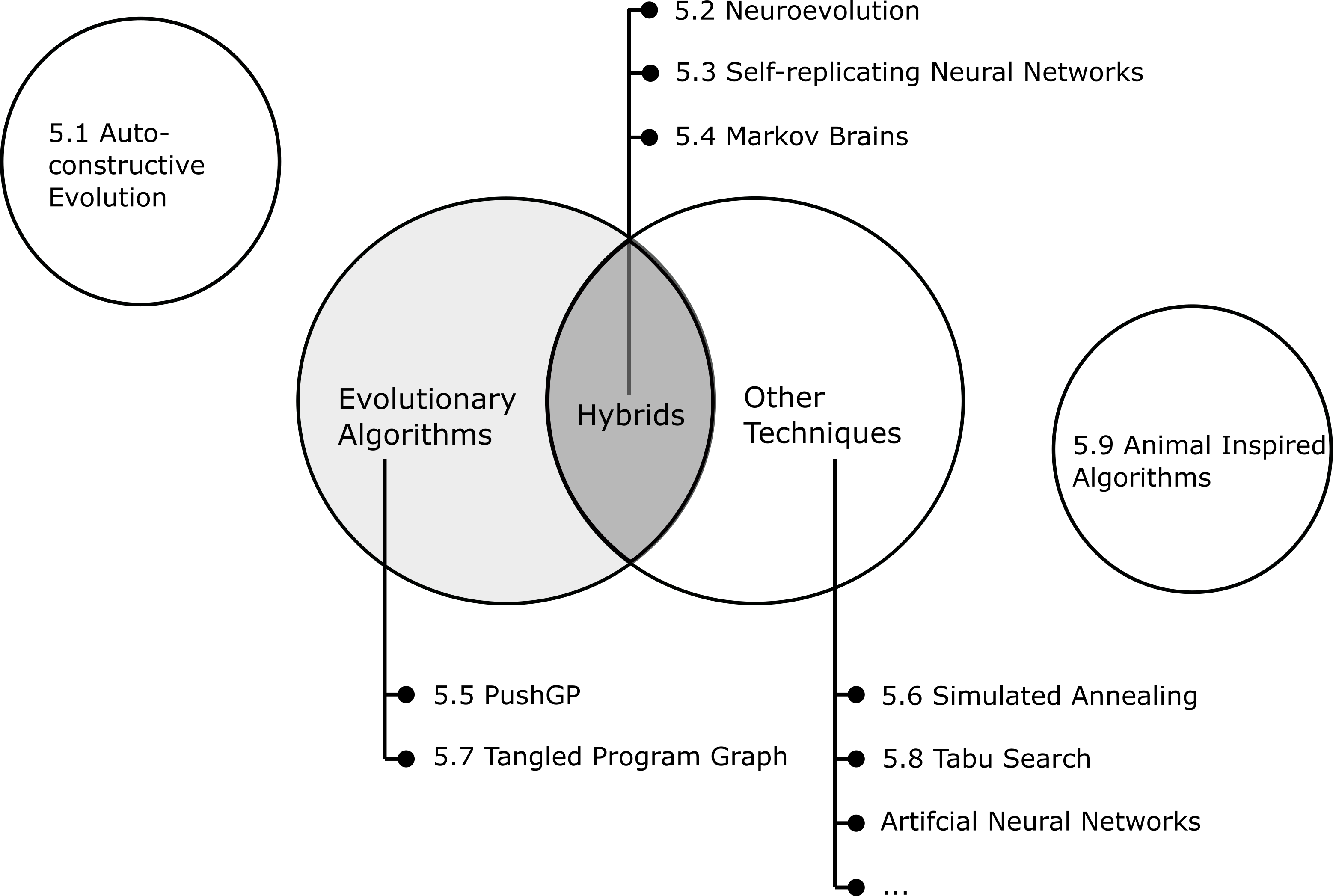}
\end{center}
\caption{Relationships between specialized techniques and concepts}
\label{figure:easpec}
\end{figure}

Figure \ref{figure:easpec} shows the relationships between the specialized techniques and concepts. These relationships are more tenuous than the relationships found between the various traditional EA techniques.  

\subsection{Auto-constructive Evolution} \label{autoconstructiveevolution}

\textit{Auto-constructive Evolution} [\ref{ref:autoconstructive}] is where instead of having an overarching algorithm orchestrating the artificial evolution process, entities themselves are given the ability to undergo evolution. This means that children are constructed by their own parents. Parents have an ability to produce children, without the need of a master synthetic algorithm. This is in contrast to the more traditional EAs where the artificial replication occurs at a higher level.\\

\textbf{What problems does auto-constructive evolution solve?} Provides a method to carry out micro-evolution. This type of evolution is more inline with the goal of building \textit{Artificial Life}. Potentially has an advantage to scale since it requires no centralized coordination. 

\subsection{Neuroevolution, or Deep Neuroevolution} \label{section:neuroevolution}

\textit{Neuroevolution} [\ref{ref:EvoAutoML},\ref{ref:uberneuroevolution},\ref{ref:automl},\ref{ref:ERL}] is classed as a hybrid solution. It has become more popular in recent years. It was a noticeably hot topic in GECCO 2018 [\ref{ref:GECCO2018}]. Neuroevolution is the concept of using some form of GA to discover the optimal way to setup a Deep Neural Network (DNN). The entities being optimized are artificial neural networks (ANNs). Or, it can be used in combination with supervised learning and reinforcement learning (RL) techniques. The family of neuroevolution algorithms can be further classified based on how the candidate solutions are encoded and how much they can vary. As mentioned previously, CGP (section \ref{cartesiangeneticprogramming}) can also be an effective way to evolve ANNs by adding weights to the links.

\begin{itemize}
\item \textbf{Direct encoding}: parameters of every artificial neuron and connection are part of the solution encoding.
\item \textbf{Indirect encoding}: is a "recipe" for generating ANNs. The topology can be either fixed or evolving. Fixed means that only the connection weights are optimized, whereas evolving means both the connection weights and the topology of the ANN are modified. This latter class of algorithms is commonly called Topology and Weight Evolving Artificial Neural Network algorithms (TWEANNs).
 
\end{itemize}

Notable examples of TWEANNs are Neuroevolution of Augmenting Topologies (NEAT) [\ref{ref:NEAT}] and its successor HyperNEAT [\ref{ref:HyperNEAT}]. The former uses a direct encoding while the latter uses an indirect encoding called "Compositional Pattern Producing Networks" (CPPN) [\ref{ref:cppn}]. Another example of a TWEANN approach using an indirect encoding is "Evolutionary Acquisition of Neural Topologies" (EANT2) [\ref{ref:EANT2}].\\

\textbf{What problems does Neuroevolution solve?} Neuroevolution combines ideas from Genetic Algorithms and Artificial Neural Networks. It can evolve both ANN weights and topologies making it an attractive alternative to ML hyper-parameter hand-crafting. By contrast with backpropagation, neuroevolution is not limited to differentiable domains [\ref{ref:ERL}]. 

\subsection{Self-replicating Neural Networks}

\textit{Self-replicating Neural Networks} [\ref{ref:quine}] is a relatively new idea where the ANNs themselves re-configure. Currently the idea is to have the network learn by producing their own weights as output also to have \textit{regeneration}. Regeneration is the concept of training an ANN by inserting predictions of its own parameters. This technique is still being researched and is at a very early stage. Vigorous exploration is still required but this idea has potential to be become more important. Expect to see forward progress in this area.\\

\textbf{What problems do Self-replicating Neural Networks solve?} Self-replication is still a new idea. Early research shows some promise in the area of continual ANN improvement using natural selection. 

\subsection{Markov Brains} \label{markovbrains}

\textit{Markov Brains} [\ref{ref:markovbrains}] is in an early stage. Markov Brains belong to the same hybrid group as neuroevolution. Based on ANNs with some significant differences. Normal ANNs are designed with layers built-up from nodes with the same functional characteristic. Markov Brains are networks built from nodes with different computational characteristics. The components interact with each other, and can connect with external sensors (physical inputs) and actuators (physical outputs).\\

\textbf{What problems do Markov Brains solve?} This is still relatively early days for Markov Brains but they are showing some early promise especially in unsupervised learning. By being a more flexible substrate than ANNs, they could also lead to a more general understanding of the role recurrence plays in learning. Looking forward to see the next papers on this subject. 

\subsection{PushGP} \label{pushgp}

\textit{PushGP} [\ref{ref:pushgp}] is a family of programming languages which have been specifically designed for evolution to be applied-to i.e. an evolution target. It is based on a stack execution model. Each datatype has a separate stack. Code is treated as a manipulated datatype. It has been subjected to continuous research over a number of years and so there are many iterations and implementations. These  variations include ones that allow for auto-constructive evolution (see section \ref{autoconstructiveevolution}).\\

\textbf{What problems does PushGP solve?} PushGP is designed to be an evolutionary target language rather than forcing a standard programming language to evolve. In other words, instead of using an existing programming language which is not evolutionary friendly, PushGP goes the other way by making it evolutionary friendly. 

\subsection{Simulated Annealing} \label{simulatedannealing}

\textit{Simulated Annealing} [\ref{ref:simon}] is not an evolutionary algorithm in itself but is frequently used in-conjunction with EAs. It is where a problems starts being exploratory and as it gets nearer to a possible solution moves more to being exploitative. Meaning that more of a stochastic approach is adopted at the beginning and as a good enough solution becomes nearer a more combinational approach is adopted (see section \ref{exploitive-exploratorycontrol}).\\ 

\textbf{What problems does Simulated Annealing solve?} Simulated Annealing is most useful when the problem-domain requires more of an exploratory approach at the beginning but as the solution becomes more in view a more exploitative approach is adopted.

\subsection{Tangled Program Graph, TPG}

Tangled Program Graph (TPG) [\ref{ref:tangledgraph}] is another relatively new technique. A method of managing programs to scale. The scale is used to handle more complicated tasks such as game playing. Provides a method to manage continuous evolution of independent and co-evolved populations. Shown to have some promising results when compared with equivalent deep learning methods. And requires significantly lower computation requirement.\\

\textbf{What problems does TPG solve?} TGP is both efficient and proven to handle complex dynamic problems such as the traditional  game playing benchmarks. This is still an early research area and as such should be monitored.

\subsection{Tabu search}

\textit{Tabu search} [\ref{ref:tabu}] is similar to Simulated Annealing (see section \ref{simulatedannealing}) in that it is often used in conjunction with EAs. Tabu search relaxes the local search method by allowing a not-so-good solution to progress-forward over solutions which have already been visited.\\

\textbf{What problems does Tabu search solve?} Tabu search solves the problem of getting off a local maxima by placing solutions which have already been visited onto a tabu list. The list is used as a method to avoid going down known previously explored search paths.

\subsection{Animal inspired algorithms}

The \textit{Animal Inspired Algorithms} [\ref{ref:simon}] are biology inspired algorithms. Never-the-less they deserve a reference within the context of EAs. There are a surprisingly large number of animal inspired algorithms. The more famous are swarm, ant and frog algorithms but the actual list is considerably longer. Each animal inspired algorithm provides some unique quality like flying, jumping or walking. They are important algorithms; swarm algorithm, for instance, can be used to control a collection of drones. Ant algorithms can be used to explore an unknown terrain for searching useful resources.\\ 
   
\textbf{What problems do Animal Inspired Algorithms solve?} This is a broad group of specialized algorithms which solve very specific problems. 

%% file: sections/domainmap.tex
\section{Problem-domain mapping}

The most important part of any algorithm is what can it accomplish. In this section we attempt to map specific problem-domains to potential techniques. We must stress that this is not exhaustive and may change dramatically between reviews. It will act as an important baseline for any future reports. 

%
%

\subsection{Specific problem-domain mappings}

Here we map the general problem-domains and the specific technique or techniques. These problem-domains are traditional problems found in industry. 

\subsubsection{Variable and parameter optimization} \label{problemdomain:parameter}

Parameter and variable optimization is a process of finding the best set of parameter values for a problem. The problems in themselves are complicated or the number of parameters is extremely large. If neither is true then a more traditional method of optimization may be a better route to a solution.\\ 

ES or GA are the normal solutions for this type of problem-domain. It has been shown that a GA can handle up to a million variables, as discussed in the Tutorial on \textit{Next Generation Genetic Algorithms} GECCO 2018 [\ref{ref:GECCO2018},\ref{ref:GECCO2018TUT}]. This makes the search space difficult or impossible for traditional methods and the only remaining true competitor is a pure stochastic method.\\

See sections,
\begin{itemize}
	\item \ref{evolutionarystrategy} Evolutionary Strategy, ES
	\item \ref{geneticalgorithms} Genetic Algorithms, GA  
\end{itemize}

\subsubsection{Symbolic and polynomial regression}

This is the problem when given a data-set finding the equivalent mathematical equation. This is a particularly popular and important activity in many industries. The requirement is to find a matching equation for the data. EAs are adopted when traditional regression methods fail. The automotive industry are actively involved in this area since they frequently need to confirm theoretical equations with practical data. The techniques help find the equation that matches the real data independent of theory.\\

GP and all subclasses can handle symbolic and polynomial regression.\\

See sections,
\begin{itemize}
	\item \ref{geneticprogramming} Genetic Programming, GP 
	\item \ref{lineargeneticprogramming} Linear Genetic Programming, LGP 
	\item \ref{cartesiangeneticprogramming} Cartesian Genetic Programming, CGP
	\item \ref{grammaticalevolution} Grammatical Evolution, GE
	\item \ref{pushgp} PushGP
\end{itemize}

\subsubsection{Automated code production} \label{automatic-code-production}

The goal is to produce new code without human involvement. Once the programming language, representation and goal have all been chosen, EAs can explore the problem-domain in search of an optimal solution. If a final code solution is found then it will fall into one of three criteria i.e. precise/accurate, "good enough" or a multi-objective compromise (along the pareto curve).\\  

Historically, EAs have mostly targeted programming languages such as LISP and low-level assembly language. Both have strict input-output formats which can simplify mutation and the joining of code segments. This avoids introducing syntax errors due to interface inconsistencies. Today the popular programming languages are Python (as a replacement for LISP) and Intermediate Representation (as a replacement for assembler instructions). Both offer new opportunities and challenges.\\ 

GP, LGP, GE, CGP, and PushGP are all techniques that produce code. Due to the fact that the code is automatically generated it is likely that the end result is difficult or unreadable by humans.\\

See sections,
\begin{itemize}
	\item \ref{geneticprogramming} Genetic Programming, GP
	\item \ref{lineargeneticprogramming} Linear Genetic Programming, LGP 
	\item \ref{grammaticalevolution} Grammatical Evolution, GE
	\item \ref{cartesiangeneticprogramming} Cartesian Genetic Programming, CGP
	\item \ref{pushgp} PushGP 
\end{itemize}

\subsubsection{Regular expression}

Automated generation of \textit{regular expression}. This can be achieved using Genetic Programming [\ref{ref:GPRE}]. Where EAs are used to explore expression strings.\\

See section,

\begin{itemize}
    \item \ref{geneticprogramming} Genetic Programming
\end{itemize}

\subsubsection{Circuit design}

\textit{Circuit design} is similar to low-level assembly instructions (see section \ref{automatic-code-production}) in that it has been successfully explored using EAs. The rules for circuit design is relatively simple and so EAs can explore the problem space relatively easily. This area was explored intensely in the 1990's but has potential to see a revival as system complexity increases. \\

CGP is particularly good at exploring circuit design since it handles spatial problems using Cartesian co-ordinates.\\

See sections,
\begin{itemize}
	\item \ref{geneticprogramming} Genetic Programming, GP
	\item \ref{cartesiangeneticprogramming} Cartesian Genetic Programming, CGP
\end{itemize}  

\subsubsection{Code improvement and optimization}

This is an up and coming area. The evolutionary process starts from a working code base, as compared with an initial random seed population. The existing working code is optimized towards a set of new objectives or is transitioned to fit within a new set of constraints. The new objectives could include specific performance features or any other similar attributes. The constraints could include new code size restrictions or power consumption limitations. The working code is basically used as a seed for the first initial population. Standard evolutionary operators are then applied to search the problem space for a potentially better solution.\\ 

As a subtopic, \textit{legacy code improvement} is about taking older existing code and finding a better alternative. This better alternative can either be a metric improvement (e.g. faster, smaller code) or higher quality  (hidden or existing anomalies are removed).\\ 

EAs in this problem-domain act as extra engineers on the project, where they might or might not produce a better answer from the original. Similar to many parts of engineering, this technique relies heavily on the quality of the goal and the associated test suites.\\

GP, GI and GE are all referenced to handle code improvement and optimization.\\

See sections,
\begin{itemize}
	\item \ref{geneticprogramming} Genetic Programming, GP
	\item \ref{geneticimprovement} Genetic Improvement, GI
	\item \ref{grammaticalevolution} Grammatical Evolution, GE 
\end{itemize}

\subsubsection{Simulator testing}

\textit{Simulation testing} is an indirect/byproduct of using EAs. It turns out EAs can be extremely good at finding simulator inconsistencies. This is because EAs explore the simulator in a different way than an engineer or scientist would explore a simulator. In fact, there are a number of historical examples where EAs are annoyingly good at discovering faults.\\
  
All the EAs are capable of pushing the limits of a simulator.

\subsubsection{Walking robot}

Robots learning to walk has been a traditionally hard problem. EAs have been involved in learning how to walk for decades. EAs have two attributes which make them particularly useful in handling these types of problems. First, EAs can learn how to improve with each evolutionary cycle (incremental improvements) and second they can adapt to changes in the environment. For example, if a physical component changes or malfunctions EAs can adapt to that change and can continue walking.\\

GAs are used extensively in Robot Walking Algorithms, for both soft and hard robots.\\

See section,
\begin{itemize}
	\item \ref{geneticalgorithms} Genetic Algorithms, GA
\end{itemize}

\subsubsection{Automated machine learning}

EAs are starting to be used in the new subject of Automated Machine Learning (or more commonly known as AutoML) [\ref{ref:automl},\ref{ref:EvoAutoML}]. AutoML is the automation of machine learning to real problem-domains. The goal is to avoid the labor intensive configuration required to setup a \textit{Deep Neural Network} (DNN). This method also potentially bypasses the requirement for domain experts. Google have shown that AutoML can successfully improve existing ML systems. Google has also shown how Evolutionary AutoML can be used to improve Image Classifiers that were originally designed by humans.\\ 

This is becoming an increasingly popular subject, especially as DNNs become more complicated and larger. DNNs are more and more being adopted to solve interesting, real problems but the complexity of setup is causing a slow down in application. The quest today is applying and configuring DNN to more problems quicker and easier. EAs are increasingly being used in this area to discover novel solutions which are more efficient.\\

In research, GAs are an alternative way to configure a DNN. The GAs provide some form of novelty. For example, in a Google Brain paper [\ref{ref:EvoAutoML}] \textit{aging} the weaker entities from the population was introduced to DNN configuration. They found that GAs achieved better results, as compared with other methods, when there is limited hardware resources.\\

One of the more famous public tools in EAs to carry this is out is TPOT [\ref{TPOT}], source code available on GitHub. It is designed to be a Data Science Assistant, written in Python. The goal is to optimize machine learning pipelines using GP.\\

See sections,
\begin{itemize}
	\item \ref{geneticalgorithms} Genetic Algorithms, GA
    \item \ref{geneticprogramming} Genetic Programming, GP
    \item \ref{section:neuroevolution} Neuroevolution, or Deep Neuroevolution
\end{itemize}

%
%

\subsection{Unusual and interesting problem-domain mappings}

Here we highlight some of the more unusual problem-domain mappings that have appeared in recent articles. It is expected that these mappings will change the most between reviews. 

\subsubsection{Configuring neuromorphic computers}

Configuration of a \textit{Neuromorphic Computer} [\ref{ref:neuromorphic}] is a more unusual problem. Current techniques map Convolutional Neural Networks (CNNs) to Spiking Neural Networks (SNN). These techniques avoid the dynamic nature and complexity of SNNs. A suggestion is to use an EA to perform these mappings. EAs are used to design simple ANNs to configure the platform. This technique allows the full utilization of a complex SNN to create small networks to solve specific problems. EAs can explore the entire parameter space of the specific hardware.   

\subsubsection{Forecasting financial markets} 

GA are used by institutional quantitative traders and other areas of the financial world [\ref{ref:financialforcast}]. Traders use software packages to set parameters that are optimized using both historical data and a GA. Depending upon the problem, the optimization can vary from which parameters are being used and the associated values to only optimizing the values. Trading comes with some risk but identifying the right parameters that relate to major market turns can be critical.

\subsubsection{Predicting future city landscapes} \label{section:landscape}

The Spanish Foundation For Science and Technology [\ref{ref:cityprediction}] have used EAs to predicts the upward growth of cities. They discovered that increases in build height follows similar development as some living systems. A GA takes historical and economic data and uses it to predict the skyline of the future. The GA predicts how the skyscrapers and other buildings increase in height.

\subsubsection{Designing an optimized floor-plan}

Using EAs to design internal building floor plans. Floor plan can be complex due to building irregularities. A GA has been applied to optimize complex floor-plans [\ref{ref:floorplanning}]. The GA successfully designed office floor plans that optimized walk times and hallways. 

\subsubsection{Antenna design}

Antennas are complicated and are mostly designed by hand. This is both time-consuming and requires many resources. EAs have been "used to search the design space and automatically find novel antenna designs". In the paper entitled \textit{Automated Antenna Design with Evolutionary Algorithm} [\ref{ref:nasaantenna}], a group of NASA scientist successfully achieve designing an antenna using digital evolution. The antenna turned out to be efficient for a variety of applications. It was unique since the final design would not have been created by a human.

\subsubsection{Defect identification of electron microscopy images}

The US Department of Energy has been using a system called \textit{Multinode Evolutionary Neural Networks for Deep Learning} (MENNDL) [\ref{ref:dlelectronmicroscopy}] to identify defects in electron microscopy images. MENNDL uses NNs to find defects out of changing data. The system runs on the Oakridge Summit supercomputer [\ref{ref:summitoakridge}]. Fully utilizing all the available compute-nodes i.e. 18,000 GPUs on 3000 nodes. It analyzes millions of networks using a "scalable, parallel, asynchronous genetic algorithm augmented with a support vector machine to automatically find a superior deep learning network topology and hyper-parameter set." [\ref{ref:dlelectronmicroscopy}]. The scale of this system makes this an impressive hybrid implementation.

%% file: sections/challenges.tex
\section{Challenges}

Challenges include some personal opinions from the experience we have had navigating the subject of EAs. The points laid out below are opinions so should be debated and discussed. They are not end points.

\begin{itemize}

\item It is our observation that the community is relatively small compared with other Machine Learning communities. The size of the community determines the level of vigor that can be applied to validate a new idea or concept. In other words, there is not enough experts in the subject to vigorously prove or disprove a concept. Many ideas, even extremely clever and good ones, go unverified and unchecked by the community. This is a problem since good ideas can go missing due to a lack of support.

\item EAs have an inherent difficulty proving they are the best solution to a specific "real-world" problem. There is no automatic methods to compare algorithms that is without any bias. In other words how do we prove, without doubt, that an EA performs better at achieving a time-to-solution than a random search. This has been a consistent issue since it is extremely difficult to prove the results from an EA experiment that is obviously without bias. What compounds the difficulty is that the problem-domains targeted are in themselves inherently complex, that is why an EAs is being used in the first place.  

\item Recently a lot of work has gone into creating synthetic problem benchmarks but there is concern less work has been applied to "real world" problems with "real world" constraints. Where bench marking and consistency is undoubtedly important, especially when comparing techniques, the most important activity is always applying algorithms to real problems.

\item The community is an old community within Machine Learning, with ideas dating back to the early 1950's but many of the original drawbacks of using evolutionary techniques have been removed since modern hardware is both abundant and high performing. This means experiments which were constrained by the hardware resources at the time can now be feasible. Population size and maximum number of generations can be made considerably larger.
  
\item Biology plays important an role but EAs are not organisms. This makes crossing terms between biology and Computer Science difficult. Computer Scientists will use the  biological terms loosely for their own purposes whereas the real biological meaning is much more complicated.

\item EAs have been proven good at tackling some hard problems but they suffer from a difficulty-of-scale. In complex systems it is important to divide-and-conquer (break the problem down into smaller elements) before attempting to produce a solution. EAs at the bottom level make a lot of sense but they become more problematic as we scale-up the problem. Research work at bigger scale problems is still immature and is also limited, for the most part, with the capabilities of the current hardware available.

\item Similar to many other AI disciplines, there is a constant struggle between \textit{make} verses \textit{buy}. There is a tendency for Researchers to re-invent the technology wheel, this is in part due to the required learning curve to attain usefulness. The amount of effort to learn a new framework can be as challenging and time consuming as creating a propriety framework from first principles. This is detrimental to the discipline, as a whole, since the re-invention slows down forward progress. The caveat is that over time frameworks become overly specific to a given problem domain, so applying the same framework to a different problem can be cumbersome.

\item Modularity is a  method to handle more complex problems by breaking the problem into more manageable components. This falls under the  divide-and-conquer strategy or scientific method. EAs solve problems using a bottom-up design and as-such are inherently more difficult to modularize. 

\item EAs may be deterministic and provable, the process by how the solution was arrived at is non-determinant and if the algorithm is run again there is no guarantee the same result will be found or any result will be found. This is the opposite of some other techniques in Machine Learning.

\item EAs are provable. Proof and explainability is becoming increasingly more important. Governments are also stepping in with new regulations, for example the \textit{General Data Protection Regulation} (GDPR) [\ref{ref:GDPR}]. GDPR is a European regulation which has direct implications on Machine Learning algorithms in general. In Article 22 [\ref{ref:Article22}], of the regulation, calls for algorithmic fairness and explainability. In other words, explain how the algorithm is correct, fair and unbiased.

\item Being an old subject EAs inherently suffers from the reinvention and rediscovery of already known concepts. Keeping everyone current with what has been published is always a challenge, especially as the amount of scientific information accelerates. 

\item Even-though EAs could potentially be the next big direction for Machine Learning, the general low funding of the subject may hold back development. This is concerning since EAs are not as well-known as other Machine Learning techniques. This situation makes obtaining core funding for EA related research difficult and very much a secondary focus. The outcome is that only a few full-time researchers worldwide focus solely on these techniques. Also, the broad interdisciplinary knowledge base required is difficult to attain.   

\end{itemize}

%% file: sections/predictions.tex
\section{Predictions}

In this section we look into the future. What may happen in the next few years with EAs. Again since these are predictions they should be treated with more questions. 

\begin{itemize}

\item It is likely that in the near future we will see more cross-pollination and collaboration between Machine Learning researchers and molecular biologists, neuroscientists and evolutionary biologists. People are held back by their specializations and generally dangerous in other areas. The philosopher Paul Feyerabend argued that the most progress is made on the boundaries between subjects [\ref{ref:feyerabend}]. It is at the boundary of biology and computer science where most advancements are likely to occur.

\item Pure research on DNNs will slow down and the DNN focus will shift to engineering and the application side for the near-term. There is a strong likelihood that Machine Learning research will diversify and be more challenging. Research problems will become hybrid, involving the merging of many techniques. A potential end goal for hybrid systems is Artificial General Intelligence (AGI). 

\item \textit{Artificial General Intelligence} is a philosophical goal, or concern depending upon whom you ask. This will potentially take decades and many stages before is can be reached, if at all. One stage is the much smaller attainable goal of producing what we call \textit{Domain Specific Artificial Life} (DSAL). Bringing together many disciplines in the desire to create solutions to a specific problem. The term is a fun play on Domain Specific Architectures as advocated by some of the hardware architecture community: small artificial lifeforms to solve specific problems.

\item New Artificial Neurons (AN) will be explored and developed. There is potential for the ANs themselves to be re-examined either by vastly increasing the number of interconnects or adding interesting attributes like co-ordinates to the model. Today's ANs have limited interconnects, whereas the biological neurons have substantially more interconnects. Then evolving these models as required. These are ideas that people like Julian Miller have put forward. Whatever future direction is taken, the AN model will most likely change over the next few years. In all likelihood we will end up with many AN models to choose from. This change will occur as our understanding of \textit{Neuroscience} and biological mechanisms increases.

\item As James Shapiro points out, there are many genetic mechanisms that could be incorporated into existing and new EAs [\ref{ref:shapiro}]. One such concept is Horizontal Gene Transfer (HGT) [\ref{ref:hgt}]. HGT  becomes important to the community, as more advanced complex systems are attempted. HGT is the ability for useful genes (or code segments) to quickly transfer across species (islands). There are Computer Science implications for such ideas. As with Richard Feynman's \textit{"There's plenty of room at the bottom"}, when referring to molecular chemistry, there is plenty of room with evolutionary biology and its application to EAs. 

\item EAs can potentially be used to explore \textit{causality}. Judea Pearl [\ref{ref:cause}] has given the industry a challenge to explain Machine Learning outcomes and identify the causes of the outcomes. In particular ideas like  \textit{counterfactual}, where forward predictions about the future can be made by inserting a change or deciding not to insert a change. This involves not just providing data correlation but creating a model on how the data was created.

\item Obfuscation may allow EAs to get involved in security, privacy and data cloaking.

\item EAs are already being used in the field of \textit{Quantum Computing} (QC), and we can expect more activity in this area. Either to configure or to handle the complicated data output. 2017 saw the Humies Award [\ref{humies}] go to an Australian Team [\ref{ref:quantum}] using EAs applied to Quantum Computing.

\end{itemize}

%% file: sections/conclusion.tex
\section{Final discussion and conclusion}

The 2019 Evolutionary Algorithms Review is a baseline for future reports. We are cognizant that there are many areas which were not covered or were only covered briefly. These areas may become more important as we consider the 2020 review. Traditional EAs continue to provide useful solutions to hard problems. They are mature algorithms. They are becoming particularly useful when placed on modern hardware. The newer trends appear to indicate that hybrid solution may provide the next future capability i.e. combining evolutionary techniques with other techniques. Where EAs are used either to bring knowledge-forwarding or optimizing complex Machine Learning configurations.\\

We dedicated most of the review to the landscaping of the various techniques. This was planned to act as a baseline for future review development. For problem-domains that affect society, and subsequently industry, we introduced UCA (User Control Attributes) criteria. There are five defined attributes i.e. \textit{limiters}, \textit{explainability}, \textit{causality}, \textit{fairness} and \textit{correction}. They impose an extra level of required thinking, where the algorithms have to be both community friendly and adhere to new government regulations. The UCA will be used as a basis for a new taxonomy for EAs. Future algorithms will not only be rated on their ability to produce an outcome but on the ability to satisfy the UCA criteria in some form or other. More generally this taxonomy could be applied to other forms of Machine Learning.\\

Current thoughts on applying the UCA to EAs are as follows:

\begin{itemize}
	\item \textbf{Limiters} is the concept of restricting the capability. It is still early days but as EAs handle more problems that either affect the physical environment (i.e. control actuators), or are involved in some privacy aspect, then methods of limiting the capability or cloaking the outcomes become more important. This may require significant external technology and thought.
	
	\item \textbf{Explainability} revolves around explaining how an algorithm produces an answer. The output solution from EAs are inherently provable since they solve a specific problem but as with all algorithms of this class, explaining how the solution came about is problematic and difficult to repeat. 
	
	\item \textbf{Causality} is about moving to a higher order answer which adds the "Why" component to the answer being searched. The challenge has been set and causality will become more important as we move forward with implementing real world EAs, driven by the desire for the EAs to do more and to understand why a conclusion has been reached.
	
	\item \textbf{Fairness} is about producing an answer which is balanced and without human prejudice. This may come down to the actual input data selected and the fitness function being used. Algorithmic fairness [\ref{ref:ainow2018report}] has to be capable of detecting biases, contesting a decision and finally instigating a potential remedy. A critical method of achieving fairness is making sure that the outcome is vigorously tested. There is potential for a mathematical element of fairness to be incorporated into the evolutionary frameworks. Whereas the society aspects will remain in the human domain for the foreseeable future and may well require careful deliberate biasing to produce a fair outcome. Two features related to fairness, not covered in this review, are the broader subjects of accountability and transparency. A method of algorithmic fairness can be complicated. This is an area that requires future monitoring. Fairness may also involve studying the initial starting states, checking that the biases are not injected right at the start of the process. 
    
    \item \textbf{Correction} is about correcting a problem once it has been identified. When an error is identified/detected/contested, EAs can be assessed on how easily a remedy can be applied. EAs by definition are adaptive, and so correction can occur as part of a continuous evolutionary process (i.e. via environmental-changes) or more manually through a direct change to the fitness function and/or constraints.
\end{itemize}

Now that we have described the basics of the UCA, we will attempt in the 2020 review to apply them to the various EAs outlined in this review. This provides the new taxonomy.\\

There is a long way to go with-respect-to EAs. We are just forming a common language by which we can communicate with the biologists. This is an exciting time since the discoveries in biochemistry and synthetic biology are occurring at unprecedented rates and the capabilities of digital hardware are coming to levels that can mimic parts of the biological system. Likewise, biology can also take advantage of some of the newly gained insights from the Computer Science field.\\ 

This means that the research runway is long but at the same time we have to realize that the hardware gap in both "effective" processing and interconnects between biology and digital systems is still vast. We may have the vision to achieve biological equivalence but the current state of the hardware is both different and non-optimal for many types of problems. This is particularly interesting as we hit the End of Moore's Law and the possibility of different compute-models being introduced and forced on the industry.

\section{Acknowledgements}
We would like to acknowledge the following people for their encouragement and feedback during the writing of this review, namely Mbou Eyole, Casey Axe, Paul Gleichauf, Gary Carpenter, Andy Loats, Rene De Jong, Charlotte Christopherson, Leonard Mosescu, Vasileios Laganakos, Julian Miller, David Ha, Bill Worzel, William B. Langdon, Daniel Simon, Emre Ozer, Arthur Kordon, Hannah Peeler and Stuart W. Card.

%% file: sections/feedback.tex
\section{Feedback}

If you have any feedback or suggestions on this review please send them to andrew.sloss@arm.com with the subject line of "2019 Review Feedback" and we will consider the enhancements.

%% file: sections/references.tex
\section{References}

\begin{enumerate}

\item \label{ref:LGP} W. Banzhaf, M. Brameier, 'Linear Genetic Programming Published' by Springer, 2007

\item \label{ref:CGP} J Miller, 'Cartesian Genetic Programming', Published by Springer, 2011

\item \label{ref:cause} J. Pearl, D. Mackenzie, 'The Book Of Why : The New Science Of Cause and Effect', Published by Basic Books, 2018

\item \label{ref:simon} D. Simon, 'Evolutionary Optimization Algorithms : Biologically Inspired and Population-Based Approaches to Computer Intelligence', Published by Wiley, 2013
 
\item \label{ref:darwin} C. Darwin, 'Origin of the Species', published November 24 1859

\item \label{ref:ponyge2} M. Fenton, J. McDermott, D. Fagan, S. Forstenlechner, E. Hemberg, M. O'Neill, April 26 2017, 'PonyGE2: Grammatical Evolution in Python', \textless\url{https://arxiv.org/abs/1703.08535}\textgreater, [accessed December 28 2018]

\item \label{ref:quine} O. Chang, H. Lipson, 'Neural Network Quine', arXiv.org, May 24 2018, \textless\url{https://arxiv.org/abs/1803.05859}\textgreater, [accessed November 11 2018]

\item \label{ref:SpiNNaker} S. Furber 'SpiNNaker', Manchester University, UK, \textless\url{http://apt.cs.manchester.ac.uk/projects/SpiNNaker/project/}\textgreater, [accessed November 11 2018] 

\item \label{ref:NEAT} K. O. Stanley, R. Miikkulainen, 'Evolving Neural Networks through Augmenting Topologies' Summer 2002, \textless\url{https://dl.acm.org/citation.cfm?id=638554} [accessed April 29 2019]

\item \label{ref:HyperNEAT}  K. O. Stanley, D. B. D'Ambrosio, J. Gauci, 'A Hypercube-Based Encoding for Evolving Large-Scale Neural Networks', April 2009, \textless\url{https://ieeexplore.ieee.org/document/6792316}, [accessed April 29 2019]

\item \label{ref:hgt} R. Jain, M. C. Rivera, J. A. Lake, 'Horizontal gene transfer among genomes: The complexity hypothesis', March 1999, \textless\url{https://www.pnas.org/content/pnas/96/7/3801.full.pdf} [accessed April 29 2019]

\item \label{baldwin} T. J. H. Morgan, T. L. Griffiths, 'What the Baldwin Effect affects', 2005, \textless\url{http://cocosci.princeton.edu/papers/BaldwinEffectAffects.pdf}\textgreater, [accessed April 29 2019]

\item \label{lamarckism} R. W. Burkhardt, 'Lamarck, Evolution, and the Inheritance of Acquired Characters', August 2013, \textless\url{https://www.ncbi.nlm.nih.gov/pmc/articles/PMC3730912/}\textgreater, [accessed April 29 2019]

\item \label{ref:hyperheuristics} E. Ozcan, B. Bilgin, E. E. Korkmaz, 'A comprehensive analysis of hyper-heuristics', February 2008, \textless\url{https://www.researchgate.net/publication/220571729_A_comprehensive_analysis_of_hyper-heuristics}\textgreater, [accessed April 29 2019]

\item \label{cma-es} N. Hansen, 'The CMA Evolution Strategy: A Tutorial', April 4 2016, \textless\url{https://arxiv.org/pdf/1604.00772.pdf}\textgreater, [accessed April 29 2019]

\item \label{BNF} J. W. Backus, 'The Syntax and Semantics of the Proposed International Algebraic Language of the Zurich ACM-GAMM Conference', 1959, \textless\url{http://www.softwarepreservation.org/projects/ALGOL/paper/Backus-Syntax_and_Semantics_of_Proposed_IAL.pdf}\textgreater, [accessed April 29 2019]

\item \label{ref:gi} J. Petke, S. O. Haraldsson, M. Harman, W. B. Langdon, D. R. White, J. R. Woodward, 'Genetic Improvement of Software: A Comprehensive Survey', April 25 2017, \textgreater\url{https://ieeexplore.ieee.org/abstract/document/7911210}\textless, [access April 29 2019]

\item \label{ref:overfitting} J. Brownlee, 'Overfitting and Underfitting With Machine Learning Algorithms', March 21 2016, \textless\url{https://machinelearningmastery.com/overfitting-and-underfitting-with-machine-learning-algorithms/}\textgreater, [accessed April 29 2019]

\item \label{humies} GECCO, 'Annual "Humies" Awards For Human-Competitive Results', \textless\url{http://www.human-competitive.org}\textgreater, [accessed November 12 2018]

\item \label{ref:fieldguide} W. B. Langdon, N. F. McFree, 'A Field Guide to Geentic Programming', \textless\url{http://www.gp-field-guide.org.uk}\textgreater, [accessed December 28 2018]

\item \label{ref:kovo1} J.R. Koza,  'Genetic Programming: On the Programming of Computers by Means of Natural Selection (Complex Adaptive Systems)', Published by A Bradford Book, December 11 1992,

\item \label{ref:kovo2} J.R. Koza, 'Genetic Programming: On the Programming of Computers by Means of Natural Selection (Complex Adaptive Systems)', Published by A Bradford Book, May 17 1994

\item \label{ref:kovo3} J.R. Koza, 'Genetic Programming III: Darwinian Invention and Problem Solving (Vol 3)', Published by Morgan Kaufmann, May 14 1999 

\item \label{ref:goldberg} D.E. Goldberg, 'Genetic Algorithms in Search, Optimization, and Machine Learning', Published by Addison-Wesley Professional, January 11 1989,

\item \label{ref:autoconstructive} L. Spector, N. F. McPhee, T. Helmuth, M. M. Casale, J. Oks, 'Evolution Evolves with Autoconstruction', July 2016, \textless\url{http://faculty.hampshire.edu/lspector/pubs/wk1202-spectorA.pdf}\textgreater, [accessed December 28 2018]

\item \label{ref:GDPR} Intersoft Consulting, 'General Data Protection Regulation GDPR', \textless\url{https://gdpr-info.eu}\textgreater, [accessed November 22, 2018]

\item \label{ref:Article22} Intersoft Consulting, 'Art. 22 GDPR Automated individual decision-making, including profiling', \textless\url{https://gdpr-info.eu/art-22-gdpr/}\textgreater, [accessed November 22, 2018]

\item \label{ref:cgp_atari} D. G Wilson, S. Cussat-Blanc, H. Luga, J. F. Miller, 'Evolving simple programs for playing Atari games', June 14 2018, \textless\url{https://arxiv.org/abs/1806.05695}\textgreater, [accessed December 28 2018]

\item \label{ref:pushgp} L. Spector, 'Push, PushGP and Pushpop', \textless\url{http://faculty.hampshire.edu/lspector/push.html}\textgreater, [accessed December 28 2018]

\item \label{ref:uberneuroevolution} Uber Engineering, 'Welcoming the Era Deep Neuroevolution', \textless\url{https://eng.uber.com/deep-neuroevolution/}\textgreater, [accessed December 28 2018]

\item \label{ref:GECCO2018} ACM, 'GECCO 2018',  \textless\url{http://gecco-2018.sigevo.org/index.html/tiki-index.php?page=HomePage}\textgreater,July 2018, [accessed December 2 2018]

\item \label{ref:markovbrains} A. Hintze, J. A. Edlund, R. S. Olson, D. B. Knoester, J. Schossau, L. Albantakis, A. Tehrani-Saleh, P. Kvam, L. Sheneman, H. Goldsby, C. Bohm, C. Adami, 'Markov Brains: A Technical Introduction', September 17 2017,\textless\url{https://arxiv.org/abs/1709.05601} [accessed December 28 2018]

\item \label{TPOT} EpistasisLab, 'TPOT', Last changed August 30 2018, \textless\url{ https://github.com/EpistasisLab/tpot}\textgreater, [accessed December 28 2018]

\item \label{ref:neuromorphic} S. Buckley, A. N. McCaughan, J. Chiles, R. P. Mirin, S. W. Nam, J. M. Shainline, G. Bruer, J. S. Plank, C. D. Schuman, 'Design of superconducting optoelectronic networks for neuromorphic computing', November 2018, \textless\url{http://neuromorphic.eecs.utk.edu/raw/files/publications/2018-Buckley.pdf}\textgreater, [accessed December 28 2018]

\item \label{ref:quantum} R. Harper, R. J. Chapman, C. Ferrie, C. Granade, R. Kueng, D. Naoumenko, S. T. Flammia, A. Peruzzo,  'Explaining quantum correlations through evolution of causal models', July 2017, \textless\url{http://www.human-competitive.org/sites/default/files/harper-02-slides.pdf}\textgreater, [accessed December 28 2018]
 
\item \label{ref:financialforcast} J. Kuepper, 'Using Genetic Algorithms to Forecast Financial Market', April 24 2018, \textless\url{https://www.investopedia.com/articles/financial-theory/11/using-genetic-algorithms-forecast-financial-markets.asp}\textgreater, [accessed November 17 2018]

\item \label{ref:cityprediction} Spain Foundation for Science and Technology, 'A genetic algorithm predicts the vertical growth of cities', May 25 2018, \textless\url{https://www.eurekalert.org/pub_releases/2018-05/f-sf-aga052518.php}\textgreater,  [accessed November 17, 2018]

\item \label{ref:nasaantenna} G. S. Hornby,  A. Globus, D. S. Linden, J. D. Lohn, 'Automated Antenna Design with Evolutionary Algorithms', \textless\url{http://alglobus.net/NASAwork/papers/Space2006Antenna.pdf}\textgreater, [accessed December 2018]

\item \label{ref:alanturing} A. M. Turing, 'The Chemical Basis of Morphogenesis', August 14 1952, \textless\url{https://www.jstor.org/stable/92463?seq=1#page_scan_tab_contents}\textgreater, [access April 29 2019]

\item \label{ref:johnvonneumann} J. V. Neumann, A. W. Burks, 'Theory of Self-Reproducing Automata', Published by University of Illinois Press; First Edition edition 1966 

\item \label{ref:SNN} W. Maass, 'Networks of Spiking Neurons: The Third Generation of Neural Network Models', March 27 1996, \textless\url{https://igi-web.tugraz.at/PDF/85a.pdf}\textgreater, [accessed April 29 2019]

\item \label{ref:seanluke} S. Luke, 'Essentials of Metaheuristics', Lulu 2nd Edition, 2013, \textless\url{https://cs.gmu.edu/~sean/book/metaheuristics/}\textgreater [accessed December 28 2018]

\item \label{ref:gep} C. Ferreira, 'Gene Expression Programming: Mathematical Modeling by an Artificial Intelligence', Publishing by Springer; 2nd edition (July 11, 2006)

\item \label{ref:jeffbezos} K. Leswing, 'Jeff Bezos just perfectly summed up what you need to know about artificial intelligence', \textless\url{https://www.businessinsider.com/jeff-bezos-shareholder-letter-on-ai-and-machine-learning-2017-4}\textgreater, [accessed December 20 2018]

\item \label{ref:ainow2018report} M. Whittaker, K. Crawford, R. Dobbe, G. Fried, E. Kaziunas, V. Mathur, S. M. West, R. Richardson, J. Schultz, 'AI Now 2018 Report', \textless\url{https://ainowinstitute.org/AI\_Now\_2018\_Report.pdf}\textgreater, [accessed December 20 2018]

\item \label{ref:bertrandrussell} B. Russell, 'Value of Philosophy', \textless\url{https://web.ics.purdue.edu/\textasciitilde drkelly/RussellValuePhilosophy1912.pdf}\textgreater, [accessed December 20 2018]

\item \label{ref:automl} E. Real, 'Using Evolutionary AutoML to Discover Neural Network Architectures', Published March 15 2018, \textless\url{https://ai.googleblog.com/2018/03/using-evolutionary-automl-to-discover.html}\textgreater, [accessed December 22 2018]

\item \label{ref:esrandom} N. Maheswaranathan, L. Metz, G. Tucker, D. Choi, J. Sohl-Dickstein, 'Guided evolutionary strategies: escaping the curse of dimensionality in random search', Published December 19 2018, \textless\url{https://arxiv.org/abs/1806.10230}\textgreater, [accessed December 22 2018]

\item \label{ref:floorplanning} A. Tokmakova, 'Optimizing floorplans via experimental algorithms', December 21 2018, \textless\url{https://archinect.com/news/article/150108746/optimizing-floorplans-via-experimental-algorithms}\textgreater, [accessed December 22 2018]

\item \label{ref:gepferreira} C. Ferreira, 'Gene Expression Programming: A New Adaptive Algorithm for Solving Problems', Published 2001, '\url{https://arxiv.org/pdf/cs/0102027.pdf}', [accessed 22 December 2018]

\item \label{ref:isotrustworthiness} W. Diab, 'About JTC 1/SC 42 Artificial intelligence', Published May 30, 2018, \textless\url{https://jtc1info.org/jtc1-press-committee-info-about-jtc-1-sc-42/}\textgreater, [accessed December 22 2018]

\item \label{ref:tangledgraph} S. Kelly, M. I. Heywood, 'Emergent Tangled Graph Representations for Atari Game Playing Agents', \textless\url{https://www.researchgate.net/publication/315066110_Emergent_Tangled_Graph_Representations_for_Atari_Game_Playing_Agents}\textgreater, [accessed December 23 2018]

\item \label{ref:EvoAutoML} E. Real, A. Aggarwal, Y. Huang, Quoc V. Le, 'Regularized Evolution for Image Classifier Architecture Search', October 26 2018, \textless\url{https://arxiv.org/abs/1802.01548}\textgreater, [accessed December 28 2018]

\item \label{ref:dlelectronmicroscopy} US Department of Energy, 'Deep learning for electron microscopy', \textless\url{https://m.phys.org/news/2018-12-deep-electron-microscopy.html}\textgreater, [accessed December 2018]

\item \label{ref:summitoakridge} U.S. Department of Energy, 'ORNL Launches Summit Supercomputer', June 8 2018, \textless\url{https://www.ornl.gov/news/ornl-launches-summit-supercomputer}\textgreater, [accessed April 29 2019]

\item \label{ref:LLVM} 'The LLVM Compiler Infrastructure Project', \textless\url{https://llvm.org/}\textgreater, [accessed January 2 2019]

\item \label{ref:cppn} K. O. Stanley, 'Compositional Pattern Producing Networks: A Novel Abstraction of Development', Springer 2007, \textless\url{https://eplex.cs.ucf.edu/papers/stanley_gpem07.pdf}\textgreater, [accessed January 2 2019]

\item \label{ref:EANT2} N. T. Siebel, 'Evolutionary Reinforcement Learning', \textless\url{http://www.siebel-research.de/evolutionary_learning/}\textgreater, [accessed Jnuary 2 2019]

\item \label{ref:ERL} F. P. Such, V. Madhavan, E. Conti, J. Lehman, K. O. Stanley, J. Clune, 'Deep Neuroevolution: Genetic Algorithms are a Competitive Alternative for Training Deep Neural Networks for Reinforcement Learning', April 20 2018, \textless\url{https://arxiv.org/pdf/1712.06567.pdf}\textgreater, [accessed January 2 2019]

\item \label{ref:GPRE} M. Gibbs, 'Genetic programming meets regular expressions', August 2 2015, \textless\url{https://www.networkworld.com/article/2955126/software/genetic-programming-meets-regular-expressions.html}\textgreater, [accessed January 2 2019]

\item \label{ref:xeoml} T. Simonite, 'Moore’s Law Is Dead. Now What?',  May 13 2016, \textless\url{https://www.technologyreview.com/s/601441/moores-law-is-dead-now-what/}, [accessed January 7 2019]

\item \label{ref:xeoml2}  B. Bailey, 'The Impact Of Moore’s Law Ending', October 29 2018, \textless\url{https://cacm.acm.org/news/232532-the-impact-of-moores-law-ending/fulltext}\textgreater, [accessed April 29 2019]

\item \label{ref:post3nm} N. Dahad, 'Imec, ASML Team on Post-3nm Lithography', October 24 2018, \textless\url{https://www.eetimes.com/document.asp?doc_id=1333896}\textgreater, [accessed January 7 2019]

\item \label{ref:googlefairness} A. Zaldivar, 'Introduction to Fairness in Machine Learning', November 27 2018, \textless\url{https://developers.googleblog.com/2018/11/introduction-to-fairness-in-machine.html}\textgreater, [accessed January 8 2019]

\item \label{ref:guardrails} Y. Sverdlik, 'Google is Switching to a Self-Driving Data Center Management System', August 02 2018, \textless\url{https://www.datacenterknowledge.com/google-alphabet/google-switching-self-driving-data-center-management-system}\textgreater, [accessed January 8 2019]

\item \label{ref:tabu} J. Brownlee, 'Tabu Search', 2015, \textless\url{http://www.cleveralgorithms.com/nature-inspired/stochastic/tabu_search.html}\textgreater, [accessed January 10 2019]

\item \label{ref:selfassmebly}  N. Krasnogor, S. Gustafson, D. A. Pelta, J. L. Verdegay,'Systems Self-Assembly, Volume 5: Multidisciplinary Snapshots (Studies in Multidisciplinarity) 1st Edition',  Published by Elsevier Science, May 12th 2008

\item \label{ref:drift1} J. Arnold, 'Genetic Drift', 2001 \textless\url{https://www.sciencedirect.com/topics/neuroscience/genetic-drift}\textgreater, [accessed April 29 2019]

\item \label{ref:drift2} Understanding Evolution, 'Genetic drift', unknown date, \textless\url{https://evolution.berkeley.edu/evolibrary/article/evo_24}\textgreater, [accessed February 4 2019]

\item \label{ref:island} M. Ammi, S. Chikhi, 'A Generalized Island Model Based on Parallel and Cooperating Metaheuristics for Effective Large Capacitated Vehicle Routing Problem Solving', 2015, \textless\url{https://pdfs.semanticscholar.org/9a62/be35edffa02ae2170319d002e93b5c1d9cf7.pdf?_ga=2.104627072.1852817033.1549303865-1734504371.1549303865}\textgreater, Published in the Journal of Computing and Information Technology 

\item \label{ref:framework-darwin} L. Mosescu, 'Darwin Neuroevolution Framework', 2018, \textless\url{https://github.com/tlemo/darwin}\textgreater, [accessed February 4 2019]

\item \label{ref:framework-ga} M. Wall, 'GAlib: Matthew's C++ Genetic Algorithms Library', March 23 1996, \textless\url{http://lancet.mit.edu/galib-2.4/}\textgreater, [accessed February 4 2019]

\item \label{ref:cmsa-es} H. Beyer, B. Sendhoff, 'Covariance Matrix Adaptation Revisited – The CMSA Evolution Strategy', September 2008, \textless\url{https://www.researchgate.net/publication/220701715_Covariance_Matrix_Adaptation_Revisited_-_The_CMSA_Evolution_Strategy_-}\textgreater, [accessed February 6 2019]

\item \label{ref:nfl} D. H. Wolpert, W. G. Macready, 'No Free Lunch Theorems for Optimization',  April 1997, \textless\url{https://ieeexplore.ieee.org/document/585893}\textgreater, [accessed April 29 2019]

\item \label{ref:islands2} Ivan, 'Parallel and distributed genetic algorithms', Mar 15 2018, \textless\url{https://towardsdatascience.com/parallel-and-distributed-genetic-algorithms-1ed2e76866e3}\textgreater, [accessed February 7 2019]

\item \label{ref:genislands} D. Izzo, M. Ruciński, F. Biscani, 'The Generalized Island Model', January 2012
\textless\url{https://www.researchgate.net/publication/285622237_The_Generalized_Island_Model}\textgreater, [accessed February 7 2019]

\item \label{ref:cause2} A. N. Sloss, 'Book Review: “The Book of Why: The New Science of Cause and Effect” By Judea Pearl and Dana Mackenzie', September 27 2019, \textless\url{https://www.linkedin.com/pulse/book-review-why-new-science-cause-effect-judea-pearl-dana-sloss/}\textgreater, [accessed February 11 2019]

\item \label{ref:dennard} R. H. Dennard, F. Gaensslen, H. N.Yu, L. Rideout, E. Bassous, A. LeBlanc, 'Design of Ion-Implanted MOSFET’s with Very Small Physical Dimensions', October 1974, \textless\url{https://www.ece.ucsb.edu/courses/ECE225/225_W07Banerjee/reference/Dennard.pdf}\textgreater, [accessed March 29 2019]

\item \label{ref:hinton} Y. LeCun, Y. Bengio, G. Hinton, 'Deep Learning', May 15 2015,
\textless\url{https://www.nature.com/articles/nature14539}\textgreater, [accessed March 29]

\item \label{ref:GECCO2018TUT} D. Whitley, 'Next Generation Genetic Algorithms', July 2018, \textless\url{http://gecco-2018.sigevo.org/index.html/tiki-index.php?page=Tutorials}\textgreater, [accessed March 2019]

\item \label{ref:feyerabend} W. J. Broad, 'Paul Feyerabend: Science and the Anarchist', November 2 1979, \textless\url{https://www.jstor.org/stable/1749231?seq=1#page_scan_tab_contents}\textgreater, [accessed April 4 2019]

\item \label{ref:shapiro} J. A. Shapiro, '21st century view of evolution: genome system architecture, repetitive DNA, and natural genetic engineering', January 4 2005, \textless\url{http://shapiro.bsd.uchicago.edu/Shapiro.2005.Gene.pdf}\textgreater, [accessed April 4 2019]

\item \label{ref:prejudice} D. Cossins, 'Discriminating algorithms: 5 times AI showed prejudice', April 27 2018 \textless\url{https://www.newscientist.com/article/2166207-discriminating-algorithms-5-times-ai-showed-prejudice/}\textgreater, [accessed April 26 2019]

\item \label{ref:codebloat} A. Purohit, N. S. Choudhari, ArunaTiwari, 'Code Bloat Problem in Genetic Programming', April 2013, \textless\url{http://www.ijsrp.org/research-paper-0413/ijsrp-p1612.pdf}\textgreater [accessed May 26 2019]

\end{enumerate}